\documentclass[sigconf]{acmart}
\renewcommand\footnotetextcopyrightpermission[1]{} % removes footnote with conference information in first column
\usepackage{graphicx}
\usepackage{multirow}
\usepackage{makecell}
\usepackage{amsmath}
\usepackage{subfloat}
\usepackage{subcaption}
\usepackage{enumitem}
\AtBeginDocument{%
  }

\setcopyright{acmlicensed}
\copyrightyear{2024}
\acmYear{2024}
\acmDOI{XXXXXXX.XXXXXXX}

\acmConference[Preprint]{Preprint}{2024}{Sydney}
\acmISBN{978-1-4503-XXXX-X/18/06}

\begin{document}

%%
%% The "title" command has an optional parameter,
%% allowing the author to define a "short title" to be used in page headers.
\title{T-JEPA: A Joint-Embedding Predictive Architecture for Trajectory Similarity Computation}

%%
%% The "author" command and its associated commands are used to define
%% the authors and their affiliations.
%% Of note is the shared affiliation of the first two authors, and the
%% "authornote" and "authornotemark" commands
%% used to denote shared contribution to the research.
\author{Lihuan Li}
% \authornote{Both authors contributed equally to this research.}
% \email{trovato@corporation.com}
% \orcid{1234-5678-9012}
% \author{G.K.M. Tobin}
% \authornotemark[1]
\email{lihuan.li@student.unsw.edu.au}
\affiliation{%
  \institution{The University of New South Wales}
  \city{Sydney}
  \state{NSW}
  \country{Australia}
}

\author{Hao Xue}
\email{hao.xue1@unsw.edu.au}
\affiliation{%
  \institution{The University of New South Wales}
  \city{Sydney}
  \state{NSW}
  \country{Australia}}

\author{Yang Song}
\email{yang.song1@unsw.edu.au}
\affiliation{%
  \institution{The University of New South Wales}
  \city{Sydney}
  \state{NSW}
  \country{Australia}
}

\author{Flora Salim}
\email{flora.salim@unsw.edu.au}
\affiliation{%
 \institution{The University of New South Wales}
 \city{Sydney}
 \state{NSW}
 \country{Australia}}

%%
%% By default, the full list of authors will be used in the page
%% headers. Often, this list is too long, and will overlap
%% other information printed in the page headers. This command allows
%% the author to define a more concise list
%% of authors' names for this purpose.
\renewcommand{\shortauthors}{Li et al.}

%%
%% The abstract is a short summary of the work to be presented in the
%% article.
\begin{abstract}
Trajectory similarity computation is an essential technique for analyzing moving patterns of spatial data across various applications such as traffic management, wildlife tracking, and location-based services. Modern methods often apply deep learning techniques to approximate heuristic metrics but struggle to learn more robust and generalized representations from the vast amounts of unlabeled trajectory data. Recent approaches focus on self-supervised learning methods such as contrastive learning, which have made significant advancements in trajectory representation learning. However, contrastive learning-based methods heavily depend on manually pre-defined data augmentation schemes, limiting the diversity of generated trajectories and resulting in learning from such variations in 2D Euclidean space, which prevents capturing high-level semantic variations. To address these limitations, we propose T-JEPA, a self-supervised trajectory similarity computation method employing Joint-Embedding Predictive Architecture (JEPA) to enhance trajectory representation learning. T-JEPA samples and predicts trajectory information in representation space, enabling the model to infer the missing components of trajectories at high-level semantics without relying on domain knowledge or manual effort. Extensive experiments conducted on three urban trajectory datasets and two Foursquare datasets demonstrate the effectiveness of T-JEPA in trajectory similarity computation.
\end{abstract}

%%
%% The code below is generated by the tool at http://dl.acm.org/ccs.cfm.
%% Please copy and paste the code instead of the example below.
%%


%%
%% Keywords. The author(s) should pick words that accurately describe
%% the work being presented. Separate the keywords with commas.
%% A "teaser" image appears betweenf the author and affiliation
%% information and the body of the document, and typically spans the
%% page.
% \begin{teaserfigure}
%   \includegraphics[width=\textwidth]{sampleteaser}
%   \caption{Seattle Mariners at Spring Training, 2010.}
%   \Description{Enjoying the baseball game from the third-base
%   seats. Ichiro Suzuki preparing to bat.}
%   \label{fig:teaser}
% \end{teaserfigure}

% \received{20 February 2007}
% \received[revised]{12 March 2009}
% \received[accepted]{5 June 2009}

%%
%% This command processes the author and affiliation and title
%% information and builds the first part of the formatted document.
\maketitle

% 1. what is Architecture for Trajectory Similarity Computation
% why you study this topic, prove it is useful or important
% 2. existing solution -> drawbacks
% 3. designing your model challenges (option)
% 4. overview your model and highlights
%yesyes, noted
\section{Introduction}
With the rapidly evolving landscape of urban transportation and the emerging availability of GPS-enabled devices, vast amounts of trajectory data are generated by daily travel. Discovering the underlying moving patterns and social behaviors by modeling trajectory similarities becomes crucial for various applications, including trajectory clustering \cite{fang20212}, anomaly detection \cite{liu2020online}, route planning \cite{liu2020popular} and location-based services \cite{han2021graph,li2024large}. 

To explore trajectory similarity measurements, initial heuristic approaches \cite{chen2005robust,vlachos2002discovering,alt2009computational,alt1995computing,yi1998efficient,chen2004marriage} operate on pairwise point matching between two trajectories. These methods are commonly considered inefficient for quadratic time complexity $O(n^2)$ when matching points of trajectories where $n$ denotes the average trajectory length. Moreover, heuristic methods are based on hand-crafted rules for point matching, which struggle to adapt to varying length trajectories and accommodate diverse similarity metrics such as spatial relations and trajectory geometry \cite{fang20212,yao2022trajgat}. Hence, deep learning models are applied to learn more robust and adaptive trajectory similarity measures. Supervised learning methods like NEUTRAJ \cite{yao2019computing} and TrajGAT \cite{yao2022trajgat} encourage the models to adapt to multiple heuristic measures with the help of Recurrent Neural Networks (RNNs) \cite{hochreiter1997long}, Transformers \cite{vaswani2017attention} or Graph Neural Networks (GNNs) \cite{velickovic2017graph}. These methods aim to approximate only heuristic-measured labels, restricting the models to learn more generalized and robust representations. To extensively capture the intrinsic patterns and structures of trajectories to adapt to a variety of routes and regions, self-supervised learning methods catch the attention of trajectory similarity computation. For example, t2vec \cite{li2018deep} pioneers the develops a generative encoder-decoder architecture with a spatial proximity-aware loss function to learn consistent trajectory representations. Due to the nature of similarity computation, contrastive learning is highly beneficial in comparing data samples as it focuses on comparing data samples. Recent work TrajCL \cite{chang2023contrastive} adopts MoCo\cite{he2020momentum} along with four novel trajectory data augmentation methods and a dual-feature module, achieving state-of-the-art performance in trajectory similarity computation.

However, there are two limitations to TrajCL. \textbf{First}, it largely benefits from data augmentation schemes that require manual effort and domain knowledge. For example, one needs to create proper transformations of a trajectory that generates variants preserving the overall features while introducing challenges to the model. However, the reliance on manual augmentation settings limits the diversity of generated trajectories, potentially restricting effective representation learning in unperceived and undiscovered scenarios. \textbf{Second}, the trajectory transformations are limited to low-level, 2D Euclidean space. Representations learned from such basic geometric transformations may fail to capture disparities in intricate, high-level semantics.

\begin{figure}
  \includegraphics[width=0.45\textwidth]{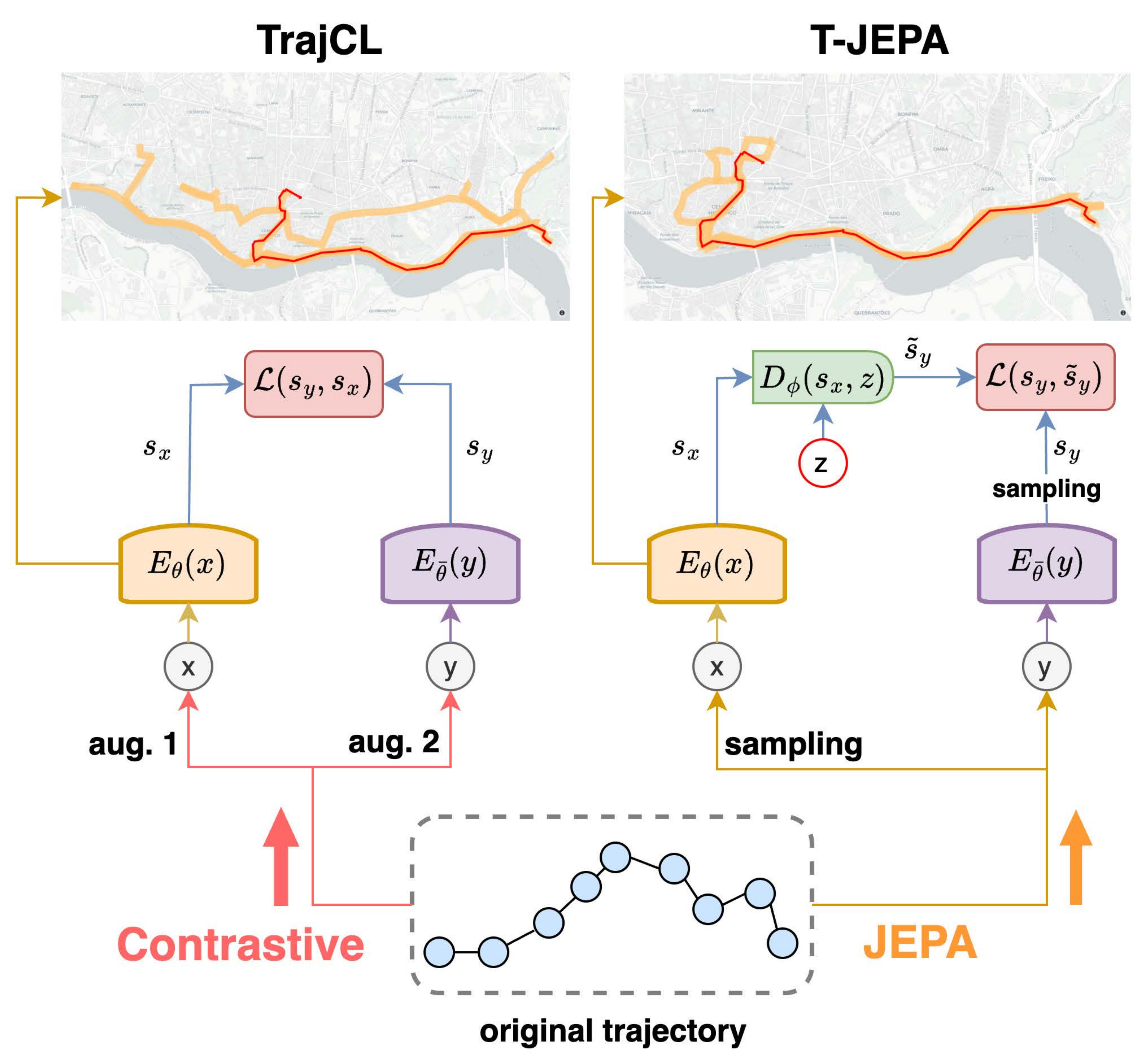}
  \caption{Given a trajectory at the bottom, we illustrate the structural differences between the contrastive learning framework (pink arrow branch) and the JEPA framework (orange arrow branch). On the top, we show the 5-NN results of TrajCL \cite{chang2023contrastive} (left figure) using contrastive learning and our proposed T-JEPA (right figure) after fine-tuning by Hausdorff measures. The query trajectory is in \textcolor{red}{red} and the matched trajectories are \textcolor{orange}{orange} heatmaps.}
  \label{fig:intro}
\end{figure}

To address these limits, we propose T-JEPA, a Trajectory similarity computation model based on Joint Embedding Predictive architecture \cite{lecun2022path}. Fig. \ref{fig:intro} briefly demonstrates the differences between the contrastive learning framework and the JEPA framework structures. T-JEPA has two branches: the target encoder branch extracts full trajectory representations and then generates multiple targets through random resampling of these representations, while the context encoder branch processes diverse context trajectories derived from an initially randomly sampled context trajectory with overlaps removed corresponding to each target. In addition to having a joint embedding structure that is similar to contrastive learning, the context encoder branch is followed by a predictive module that decodes the encoded context representations one by one to approximate the corresponding targets. 

Benefiting from this framework, T-JEPA has two advantages over existing self-supervised learning methods in trajectory similarity computation: \textbf{automated augmentations and high-level semantic understanding.} The customized automatic resampling process on trajectory representations generates diverse learning targets that avoid using any manual effort or domain knowledge such as pre-defined trajectory data augmentations \cite{chang2023contrastive}. The resampling occurs in representation space, abstracting the sampled trajectory information to a higher level and therefore creating variations beyond the 2D space. The unique predictive process maps the extracted representations to target representations instead of actual trajectory points, ensuring the encoders retain as much necessary higher-order information as possible and thereby promote high-level trajectory semantic understanding. 

Given a group of query trajectories and a group of candidate trajectories, we aim to compute similarity distances for each query-candidate pair based on their representations, where smaller distances indicate more similar pairs. This process emphasizes effective representation learning to identify the most similar matches. Consequently, both advantages of our T-JEPA can contribute to this problem by enabling more generalized and robust representations learned from the encoders. We also show an example of 5-NN visualization on the Porto dataset \footnote{https://www.kaggle.com/c/pkdd-15-predict-taxi-service-trajectory-i/data.} at the top of Fig. \ref{fig:intro} that the most similar trajectories matched by T-JEPA are more conformed to the query trajectory as it captures more accurate trajectory representations.

Moreover, we develop an $AdjFuse$ module to enrich the contextual information carried by each trajectory. We design a sliding kernel that moves along the trajectory and aggregates adjacent regional features for each point in a convolution style. The fusion of adjacent information alleviates the impact of inconsistency and discontinuity caused by low and irregularly sampled trajectories, which supports the model in maintaining invariant and robust representations from such trajectories.

We summarize our main contributions as follows:
\begin{itemize}
    \item We propose T-JEPA, the first method that applies JEPA on trajectory similarity computation. It introduces an automatic resampling process and a predictive process to augment data in abstract representation space and encourage high-level semantic understanding, which enables more generalized and robust trajectory representation learning.
    \item We design an $AdjFuse$ module with a sliding kernel to enrich the spatial contextual information from trajectories, stabilizing the learned representations from low and irregularly sampled trajectories.
    \item We experiment on both GPS trajectory datasets and FourSquare datasets to demonstrate the T-JEPA is more robust to various types of data in trajectory similarity computation.
\end{itemize}

\section{Related Work}

\subsection{Trajectory similarity computation}
Trajectory similarity computation methods are generally divided into two categories: non-learning heuristic methods and learning methods.
Heuristics methods focus on comparing the distances of point pairs from two trajectories to find the best matches with the least overall distances. EDR \cite{chen2005robust} (Edit Distance on Real Sequence) counts the number of edits needed to transform one trajectory into another in a certain threshold. LCSS \cite{vlachos2002discovering} (Longest Common Subsequence) finds the longest subsequent that appears in both trajectories with skips in the same order. Hausdorff \cite{alt2009computational} calculates the greatest mismatch between the points from two trajectories. Discret Fr\'{e}chet \cite{alt1995computing} captures the overall shape of two trajectories by considering both spatial and sequential properties. However, these methods have high time complexity as the distance computations are operated on each pair of points, causing great inefficiency. The heuristic rules also restrict their performance in measuring similarities from different perspectives.

Recent studies use deep learning methods to learn more robust and generalized trajectory features. Supervised learning methods such as NEUTRAJ \cite{yao2019computing} and Traj2SimVec \cite{zhang2020trajectory} project trajectory points on grid cells and apply LSTMs \cite{hochreiter1997long} to model trajectory sequential information. TrajGAT \cite{yao2022trajgat} represent trajectory points in hierarchical spatial regions, followed by a GAT-based \cite{velickovic2017graph} Transformer to model especially long trajectories. T3S \cite{yang2021t3s} encodes the structures and spatial information separately to improve the adaptation to various similarity measures. The objective of these supervised learning methods is to approximate the heuristic methods, resulting in the learned trajectory representations having limited generalizations. Therefore, self-supervised learning methods are leveraged to learn more robust representations. t2vec \cite{li2018deep} and CL-TSim \cite{deng2022efficient} also project trajectory points in grid cells but pre-train the cell embeddings by Skip-gram \cite{mikolov2013efficient} models to learn the spatial relationships. TrjSR \cite{cao2021accurate} converts trajectories on images and learns trajectory representations by generating super-resolution images. CSTRM \cite{liu2022cstrm} applied contrastive learning to compare both trajectory-level and point-level differences. Recent work TrajCL \cite{chang2023contrastive} develops a dual-feature attention module to effectively integrate the structural and spatial trajectory information with various data augmentation methods for contrastive learning. However, the manual data augmentation schemes in TrajCL sets a limit to the diversity of trajectory variations, especially on high-level semantics. Our self-supervised framework T-JEPA requires no such manual process and emphasizes high-level semantic understanding for trajectory representation learning, resulting in better performance in trajectory similarity computation.

\subsection{Joint-Embedding Predictive Architecture}
JEPA-based models are rapidly emerging and have been proven effective in various deep-learning tasks. It is first applied to computer vision and then to other research fields. Compared to existing self-supervised learning frameworks such as contrastive learning that relies heavily on manually defined data augmentation schemes, JEPA automates the sampling process in embedding space, followed by leveraging a novel predictive mechanism to focus on capturing complex data patterns and underlying semantics. Image-based JEPA (I-JEPA) \cite{assran2023self} first adopts this learning strategy by underscoring patch-level image semantic understanding, demonstrating better performance and efficiency than recent self-supervised learning frameworks which aim at pixel-level reconstruction \cite{caron2021emerging,zhou2021ibot}. Following this work, JEPA-based methods emerged in visual representation learning. MC-JEPA \cite{bardes2023mc} uses JEPA to integrate the visual content feature learning and optical flow estimation by establishing connections between their joint embeddings. On top of predicting image patches embedding for visual representation learning by I-JEPA, V-JEPA \cite{bardes2023v} enriches the learning process by predicting spatio-temporal region representations in video clips, effectively capturing frame changes through temporal dynamics. Besides images and videos, Point-JEPA \cite{saito2024point} is designed to benefit point cloud representation learning by accurately capturing the 3D structural proximity.

Researchers are also expanding the influence of JEPA to areas other than CV. Fei \textit{et al.} \cite{fei2023jepa} propose A-JEPA that applies JEPA to the audio and speech classification with an improved curriculum masking strategy, advancing the audio spectrogram semantic understanding. S-JEPA \cite{guetschel2024s} is the first work that uses JEPA on time series data. It excels at extracting higher-level temporal dynamics of electroencephalography (EEG) signals and proves its advancements on several brain-computer interface paradigms. In addition to the methods that apply JEPA to real-world tasks, Sobal \textit{et al.} \cite{sobal2022joint} starts from a more theoretical view and uses VICReg \cite{bardes2022variance} and SimCLR\cite{chen2020simple} to explore the robustness of JEPA on different noise changing patterns, which examines the limits and potentials of JEPA. To the best of our knowledge,
there are currently no JEPA-based methods for trajectory modeling. Our proposed T-JEPA is the first method that applies the JEPA framework to GPS trajectory representation learning with a novel AdjFuse module to identify robust and complex trajectory semantics.

\section{Preliminaries}
\textbf{Problem Definition.} A trajectory $T=\{{X_{1}, X_{2}, \ldots, X_{n}}\}$ is a sequence of GPS locations, where $n$ is the length of the trajectory and $X_{i}=(lon_{i}, lat_{i})$ is the $i$-th location represented by the longitude and latitude. Given a trajectory database $\mathcal{T}$ and each $T\in\mathcal{T}$, we aim to use a trajectory feature extractor $\mathcal{F}(T) \rightarrow h\in\mathbb{R}^d$ that maps a trajectory $T$ to an embedding vector with dimension $d$. Smaller distances $dist(\mathcal{F}(T_{i}), \mathcal{F}(T_{j}))$ of learned embedding vectors indicate more similar pairs of trajectories $(T_{i}, T_{j})$.\\
\textbf{JEPA structure}. Contrastive learning methods \cite{he2020momentum} are based on Joint Embedding Architecture (JEA) where the contrastive loss $\mathcal{L}(E_{q}(x),E_{k}(y))$ is calculated between two encoded embeddings $E_{q}(x)$ and $E_{k}(y)$. 

\begin{figure}
  \includegraphics[width=0.3\textwidth]{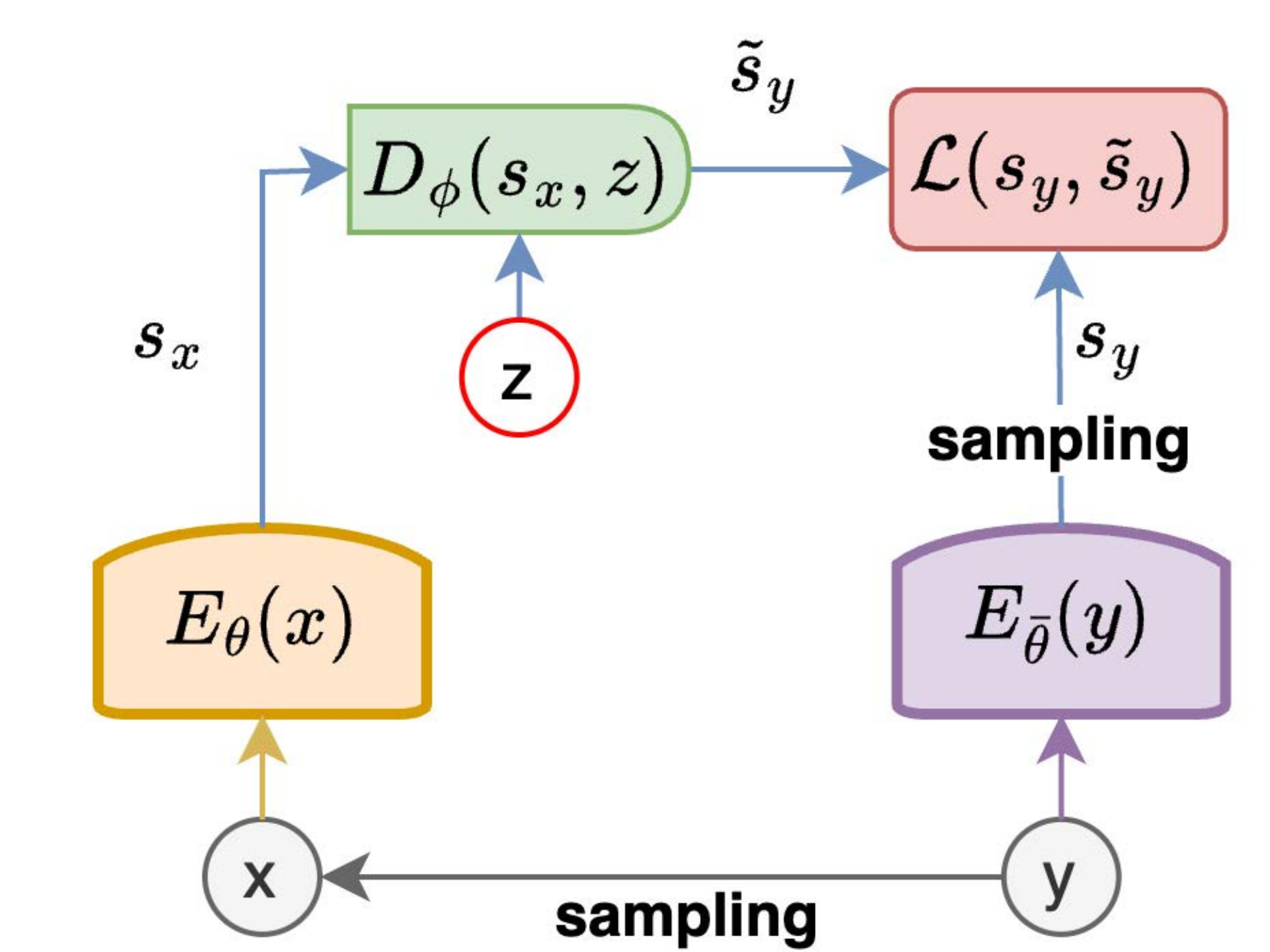}
  \caption{The structure of JEPA framework.}
  \label{fig:jepa}
\end{figure}

JEPA, as shown in Fig.\ref{fig:jepa}, has the following main components in the framework: (1) Given a data sample $y$, the target encoder $E_{\bar{\theta}}$ extracts the full data representation and generates multiple targets $s_{y}$ by resampling from the encoded representation. (2) the context encoder $E_{\theta}$ extracts the representations $s_{x}$ from the context $x$, which initially sampled a portion from full data $y$ and removed overlaps from the corresponding positions in each target. For image data, $x$ could be a set of patches of an image. (3) the predictive module $\textsf{g}_{\mathscr{\phi}}$ predicts the information $\Tilde{s}_{y}$ extracted context representations $s_{x}$ with the help of a mask token $z$ to approximate the targets $s_{y}$. Different from generative methods like Masked Autoencoders \cite{he2022masked} aiming at data-level reconstruction, the predictive module stresses predictions on the representation level, which pays attention to intrinsic dependencies of data. The mask token $z$ facilitates the predictor in capturing the necessary information from $s_{x}$. The random resampling process creates diverse subsets from data representations, ensuring the model comprehends complex relations between various portions of data.

\section{Method}
%%%% 1. existing solution limitation 
%%%% 2. motivation for the current approach
%%%% 3. what we propose, and give a preview/ overview of the technical highlights

In this section, we will elaborate on our T-JEPA in three parts: first, the model overview outlining the entire workflow; second, the main components of T-JEPA highlighting the customized design to trajectory representation learning; and third, the contextual feature enrichment process explaining how trajectory background information is strengthened by our proposed $AdjFuse$ module.

\subsection{Model Overview}

\begin{figure*}
  \includegraphics[width=.9\textwidth]{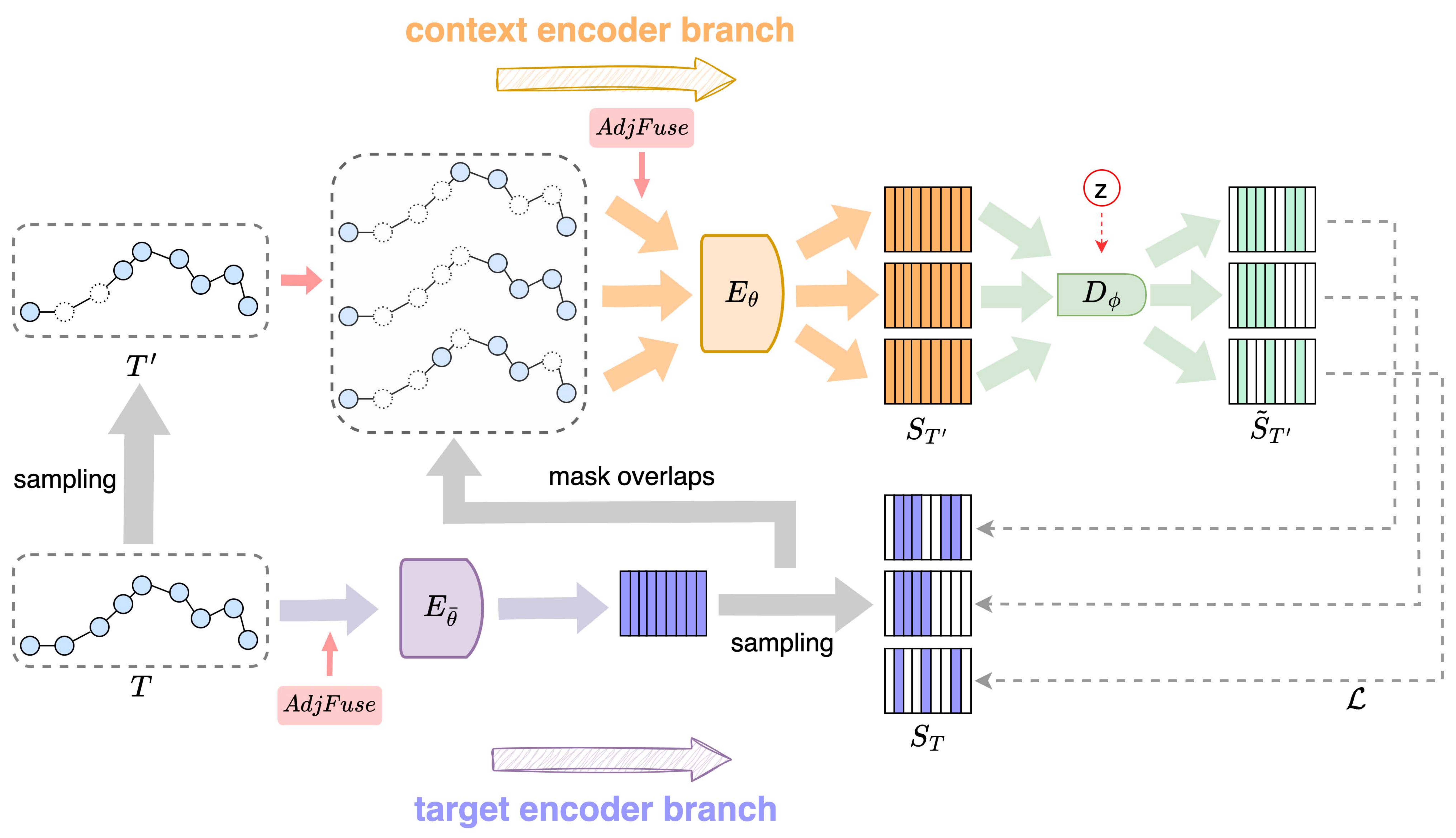}
  \caption{Our proposed T-JEPA framework. Given a trajectory, this training strategy is designed to predict multiple sampled target embeddings from one sampled context trajectory.}
  \setlength{\abovecaptionskip}{-1 cm}
  \label{fig:T-JEPA}
\end{figure*}

Our proposed T-JEPA inherits the JEPA framework but is adapted to trajectory modeling. We innovate the resampling strategy with different sampling ratios to account for varying learning difficulty levels and develop an $AdjFuse$ module to improve the robustness of learning low and irregularly sampled trajectories by aggregating the adjacent regional contextual information. Considering the dynamic nature of trajectories, we adopt one-by-one overlap removal between initially sampled context with each target. We only adopt the overall structures from JEPA as introduced in Fig. \ref{fig:jepa} as our learning framework.

The overall workflow is shown in Fig.\ref{fig:T-JEPA}. T-JEPA consists of a target encoder branch and a context encoder branch each illustrated by the purple sketch-style arrow at the bottom and the orange sketch-style arrow at the top. Given a trajectory $T$, the $AdjFuse$ module aggregates the adjacent regional features to obtain enriched trajectory contextual information. Then, the target encoder $E_{\bar{\theta}}$ extracts target features from the full trajectory $T$ and resamples these features to obtain multiple targets $S_{y}$. The sampling process is highly automated which ensures the sampled targets have a high degree of randomness. The context trajectory $T^\prime$ is initially randomly sampled from the full trajectory $T$, where overlaps are then removed corresponding to the sampled indexes of each target to generate multiple context input trajectories. It is also fed into the $AdjFuse$ module to process adjacent regional information of the sampled points only. The context encoder $E_{\theta}$ extracts the context features $S_{T^\prime}$, followed by a predictor $D_{\mathscr{\phi}}$ with mask tokens $z$ to predict the targets. This step helps capture the dependencies of the context and targets by predicting the missing information of context in the representation space. Accurate predictions indicate an effective understanding of higher-order representations of trajectories. The loss $\mathcal{L}$ is calculated between predictions $\Tilde{S}_{T}$ and the corresponding targets $S_{T}$. We apply $\text{Smooth}_{L1}$ loss to calculate the distances between each prediction and target at each corresponding sampled position.

Afterward, we apply our learned model to trajectory similarity computation. Given a trajectory $T_{i}$, we extract the representations by the learned AdjFuse module and the context encoder without any sampling process. We denote the $AdjFuse$ as a function $\mathbf{g}_{W}$ with the a set of parameters $W$ and describe the feature extraction process as:
\begin{equation}
    \mathcal{F}(T_{i}) = E_{\theta}(\mathbf{g}_{W}(T_{i}))
\end{equation}
where $\mathcal{F}(T_{i})$ can be directly compared to extracted representations $\mathcal{F}(T_{j})$ for any trajectory $T_{j}$. The backbone feature extractor $\mathcal{F}$ can also be concatenated with other models for transfer learning.

\subsection{Main Components of T-JEPA}
T-JEPA aims to predict the representations of various sampled targets from corresponding sampled contexts, with all these samples originating from a single trajectory. As shown in Fig. \ref{fig:T-JEPA}, the encoders $E_{\theta}$ and $E_{\bar{\theta}}$ have identical Transformer encoder structures but do not share weights and the predictor is a Transformer decoder. To bridge these learning components, we will elaborate on the concept of targets, context, and predictions in detail.

\noindent\textbf{Targets.} Given a trajectory $T$ with length $n$, we feed it into the AdjFuse module $\mathbf{g}_{W}$ then the target encoder $E_{\bar{\theta}}$ to extract its representations $S_{T}={S_{T1}, S_{T2}, ..., S_{Tn}}$. Next, we sample from $S_{T}$ for $M$ times with replacements to obtain the targets $S_{T}(i)=\{S_{T_{j}}\}_{j\in \mathcal{M}_{i}}$, where $S_{T}(i)$ is the $i$-th sampled target and $\mathcal{M}_{i}$ is the $i$-th sampling mask starting from a random position. Different from I-JEPA\cite{assran2023self} that samples image blocks in various aspect ratios for diverse feature learning, trajectories do not possess the same geometric property. Therefore, we design a customized sampling strategy for T-JEPA to learn from diverse sets of sampled targets across various difficulty levels. We set three target sampling ratios $\{r_{1},r_{2},r_{3}\}$ where one of them will be randomly selected for each training iteration. Additionally, we define a successive probability $p$ to determine if the samples will be taken from the trajectory representations successively or not. This approach ensures that the model is exposed to a variety of sampled targets, enhancing the robustness of the learned representations.

\noindent\textbf{Context.} The initial context trajectory $T^\prime$ is randomly sampled by a mask $\mathcal{M}_{T}$ with sampling ratio $p_{\gamma}$. Considering the dense spatio-temporal dependencies in trajectory semantics, removing overlaps from all targets for one initial context trajectory may have chances to result in very sparse information. Such situations break the basic contextual information needed to infer the entire or partial trajectory representations, which hinders effective learning. Images, on the other hand, carry sparse semantics where patch information usually contributes to the overall understanding. Therefore, we use the initially sampled context $T^\prime$ to remove overlaps with each target $S_{T}(i)$  one by one and create $M$ context inputs which are subsequently fed into the AdjFuse module $\mathbf{g}_{W}$ then the context encoder $E_{\theta}$ one by one to extract the encoded context representations $S_{T^\prime}(i)=\{S_{T^{\prime}_{j}}\}_{j\in \mathcal{M}_{T}}$.

\noindent\textbf{Predictons.} The predictor $D_{\phi}$ decodes the information from each encoded context representation jointly with mask tokens $z$, where the mask tokens are added with the positional embedding. The predictions corresponding to each target $S_{T}(i)$ conditioned by the mask tokens are denoted as $\tilde{S}_{T}(i)$.

\subsection{Contextual Feature Enrichment}
High-quality trajectory representation learning requires not only an appropriate framework but also correct and effective representations. In this subsection, we will introduce how we represent the trajectories and explain how the proposed $AdjFuse$ module can contribute to robustness. \\
\textbf{Cell representation.} We follow the previous solutions \cite{chang2023contrastive,li2018deep} and partition the study area into equally sized grid cells, where each trajectory point is assigned a cell ID. To learn the spatial relationships among cells, we use node2vec \cite{grover2016node2vec} to pre-train the cell embeddings by considering the grid as a graph $\mathcal{G}=(V,E)$ where each cell is a node $v_{i}\in V$ establishing edges $e_{i}$ to its neighboring nodes. We denote the features of node $v_{i}$ as $h_{i}$. Nearby nodes will have similar embeddings due to the spatial vicinity.

\noindent{$\textbf{AdjFuse}$.} After obtaining the pre-trained graph, we first represent each point $p_{i}=(lon_{i},lat_{i})$ in a trajectory $T$ as the node embedding $h_{\delta(p_{i})}$ where $\delta$ is a conversion function that maps the points to the corresponding node IDs. we denote the length $n$ sequence of node embeddings as $H=(h_{\delta(p_{1})},h_{\delta(p_{2})}, ...,h_{\delta(p_{n})})$ where $H\in \mathbb{R}^d$ and $d$ is the embedding dimension. 

In real cases, GPS trajectory data has noisy records \cite{li2018deep} and is usually low and irregularly sampled \cite{liang2021modeling,liang2022trajformer}. If two consecutive trajectory points are far from each other due to a long sampling interval, their corresponding node embeddings will have less correlation. Although spatio-temporal information can be compensated by powerful backbone encoders such as Transformers \cite{vaswani2017attention}, the large disparities of node embeddings still exhibit insufficient continuity of underlying routes in between. Besides, noisy records also reduce the robustness of the learned representations by disturbing the local spatio-temporal features of trajectory segments. To address this issue, previous works \cite{li2018deep,chang2023contrastive,cao2021accurate,liu2022cstrm} use data augmentations such as down-sampling or distortion to discover invariant features under the situations above. Instead, we incorporate an $AdjFuse$ module to aggregate adjacent information at each point to enrich the contextual information carried by each point.

\begin{figure}
  \includegraphics[width=0.45\textwidth]{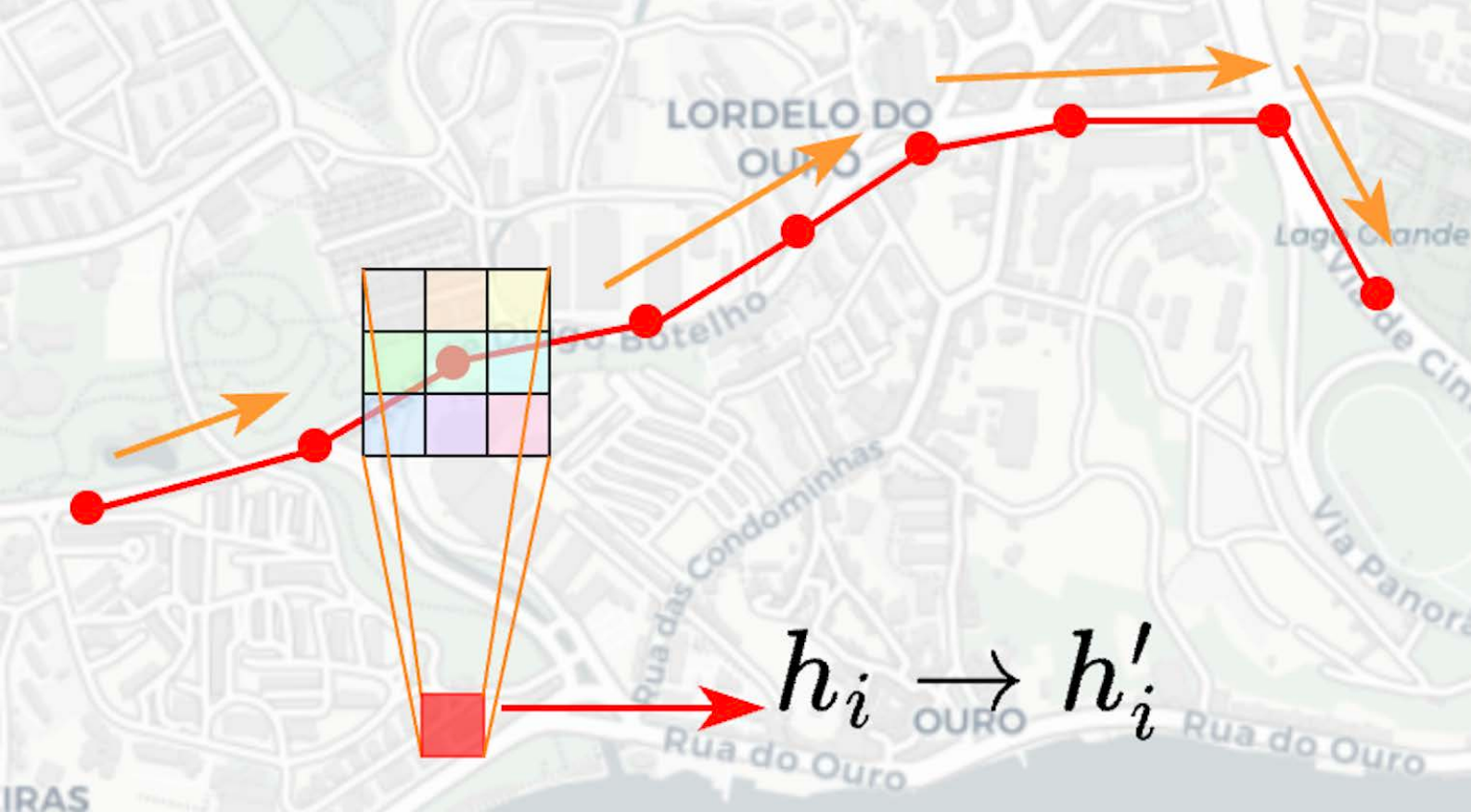}
  \caption{The illustration of the AdjFuse module.}
    \setlength{\abovecaptionskip}{-0.2 cm}
    \setlength{\belowcaptionskip}{-0.3 cm}
  \label{fig:adjfuse}
\end{figure}

Aggregating adjacent information for all nodes is infeasible as this process will take significant GPU memory. Hence, trajectory-wise operations are preferred. Specifically, after the node ID for a trajectory point $v_{i}=\delta(p_{i})$ is obtained, we further query its neighbors $\mathcal{N}(i)$ and integrate their information by the following steps:
\begin{equation}
    \setlength{\abovedisplayskip}{5pt}
    \setlength{\belowdisplayskip}{5pt}
    W^\prime = \frac{\exp(w_{j})}{\sum_{k \in (\mathcal{N}(i)\cup \{i\})} \exp(w_{k})} 
\end{equation}
where $W$ is a $3\times3$ learnable kernel covering the neighbors of node $v_{i}$ and itself. $w_{j}$ is the corresponding kernel weight currently at node $v_{j}$. We first apply softmax to obtain the normalized kernel $W^\prime$. Next, we perform the convolution-like operation:
\begin{equation}
    \setlength{\abovedisplayskip}{5pt}
    \setlength{\belowdisplayskip}{5pt}
    \tilde{h} = \sigma\left(\sum_{j \in \mathcal{N}(i)\cup \{i\}} w^{\prime}_{j} \cdot h_{j} + b\right)
\end{equation}
where $\tilde{h}$ is an intermediate node embedding, $w^{\prime}_{j}$ is the normalized weight at node $v_{j}$, and $b$ is the bias. The final node embedding of node $h_{i}$ is further obtained by:
\begin{equation}
    \setlength{\abovedisplayskip}{5pt}
    \setlength{\belowdisplayskip}{5pt}
    h^\prime_{i} = h_{i} + \tilde{W} \cdot \tilde{h}
\end{equation}
where $\tilde{W}$ conducts a linear transformation of $\tilde{h}$, followed by a residual connection to the previous node embedding $h_{i}$ to preserve useful initial node information. We illustrate the process in Fig. \ref{fig:adjfuse} to help elaborate the $AdjFuse$ module. The red line with dots is the trajectory, the $3\times3$ colored block is the weight kernel, and the orange arrows are the moving directions of the kernel. The kernel slides along the trajectory points to extract high-level local features, providing the model with richer adjacent regional contextual information. For the context encoder branch during training, the kernel is only applied to sampled points, with the kernel weights shared across both branches. The updated trajectory embeddings are denoted as $H^\prime=(h^\prime_{\delta(p_{1})},h^\prime_{\delta(p_{2})}, ...,h^\prime_{\delta(p_{n})})$ and are subsequently fed into transformer encoders $E_{\theta}$ and $E_{\bar{\theta}}$.

\section{Experiments}

\subsection{Experimental Settings}
\textbf{Datasets.} We conduct experiments on three real-world GPS trajectory datasets and two FourSquare datasets.
\begin{itemize}[left=0pt]
    \item \textbf{Porto} \footnote{https://www.kaggle.com/c/pkdd-15-predict-taxi-service-trajectory-i/data.} includes 1.7 million trajectories from 442 taxis in Porto, Portugal. The dataset was collected from July 2013 to June 2014.
    \item \textbf{T-Drive} \cite{yuan2010t,yuan2011driving} contains trajectories of 10,357 taxis in Beijing, China from Feb. 2 to Feb. 8, 2008. The average sampling interval is 3.1 minutes.
    \item \textbf{GeoLife} \cite{zheng2008understanding,zheng2009mining,zheng2010geolife} contains trajectories of 182 users in Beijing, China from April 2007 to August 2012. There are 17,6212 trajectories in total with most of them sampled in 1--5 seconds.
    \item \textbf{Foursquare-TKY} \cite{yang2014modeling} is collected for 11 months from April 2012 to February 2013 in Tokyo, Japan, with 573,703 check-ins in total.
    \item \textbf{Foursquare-NYC} \cite{yang2014modeling} is collected for 11 months from April 2012 to February 2013 in New York City, USA, with 227,428 check-ins in total.
\end{itemize}

\begin{figure}[htp]
    \centering

    % First row of figures
    \begin{subfigure}[b]{0.48\linewidth}
        \includegraphics[width=\linewidth, height=4cm]{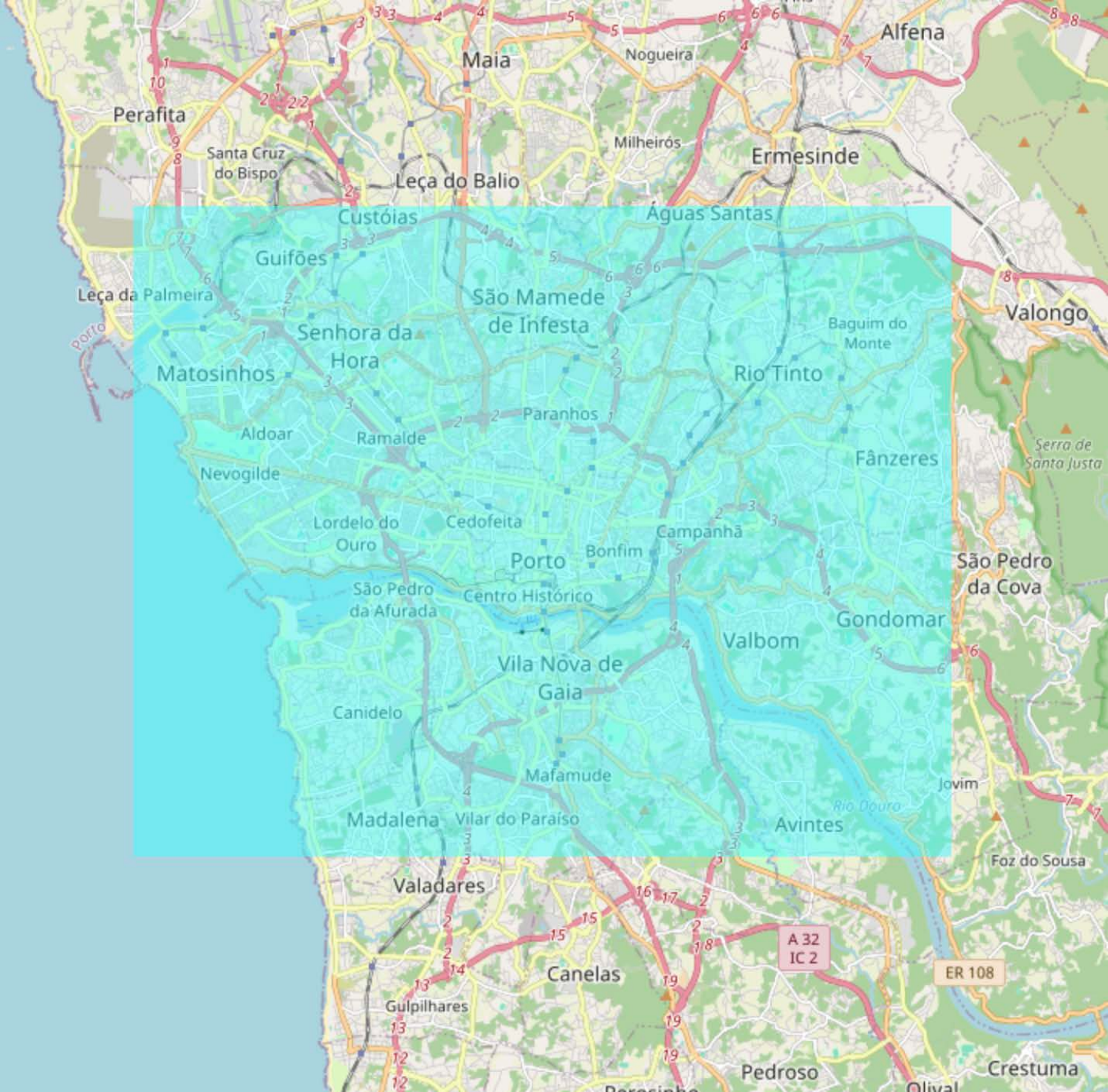}
        \caption{Porto, Portugal (183.13 $km^2$)}
        \label{fig:sub1}
    \end{subfigure}
    \hfill
    \begin{subfigure}[b]{0.48\linewidth}
        \includegraphics[width=\linewidth, height=4cm]{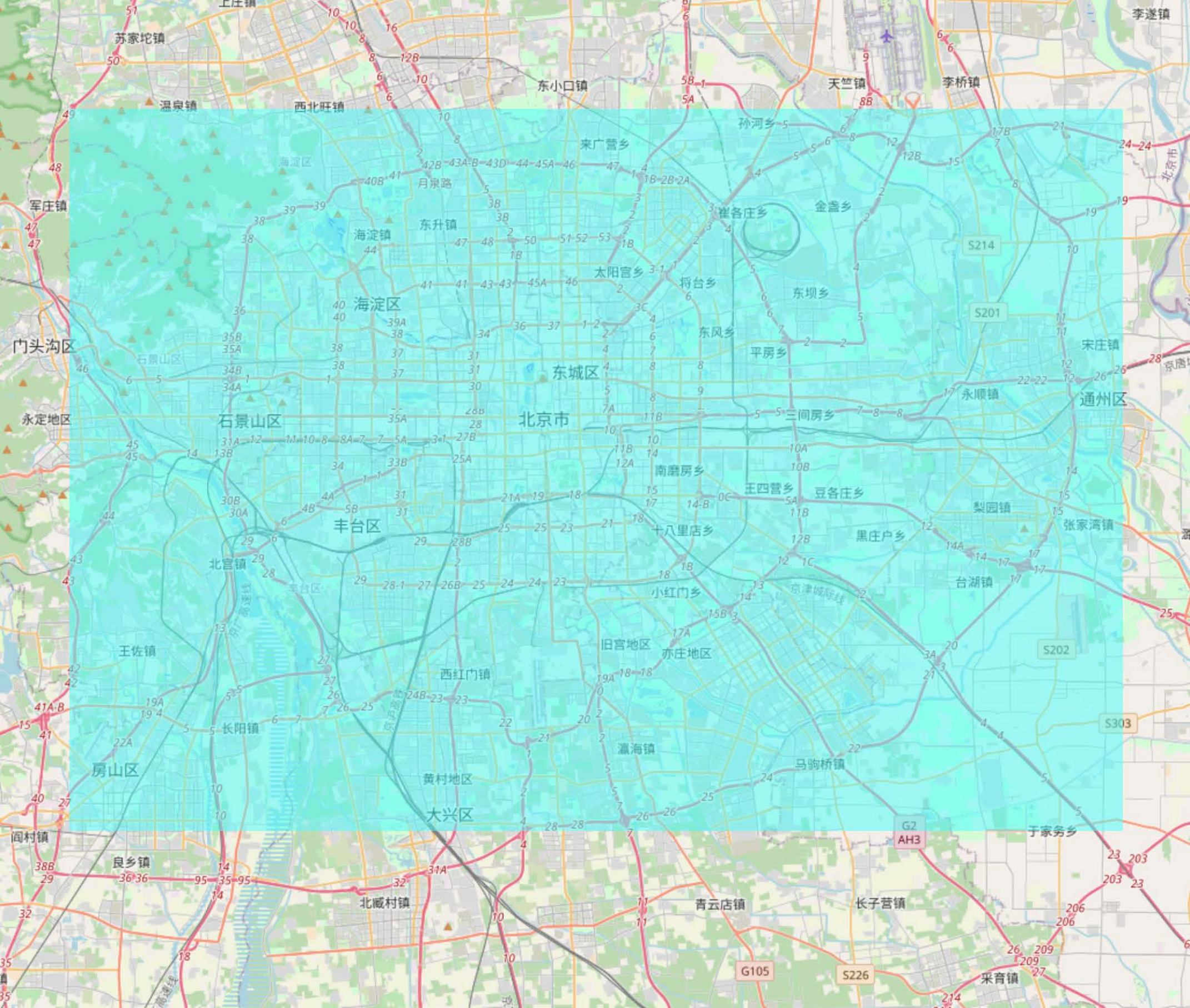}
        \caption{Beijing, China (1949.26 $km^2$)}
        \label{fig:sub2}
    \end{subfigure}
    \vspace{0.1cm} % Space between rows

    % Second row of figures
    \begin{subfigure}[b]{0.48\linewidth}
        \includegraphics[width=\linewidth, height=4cm]{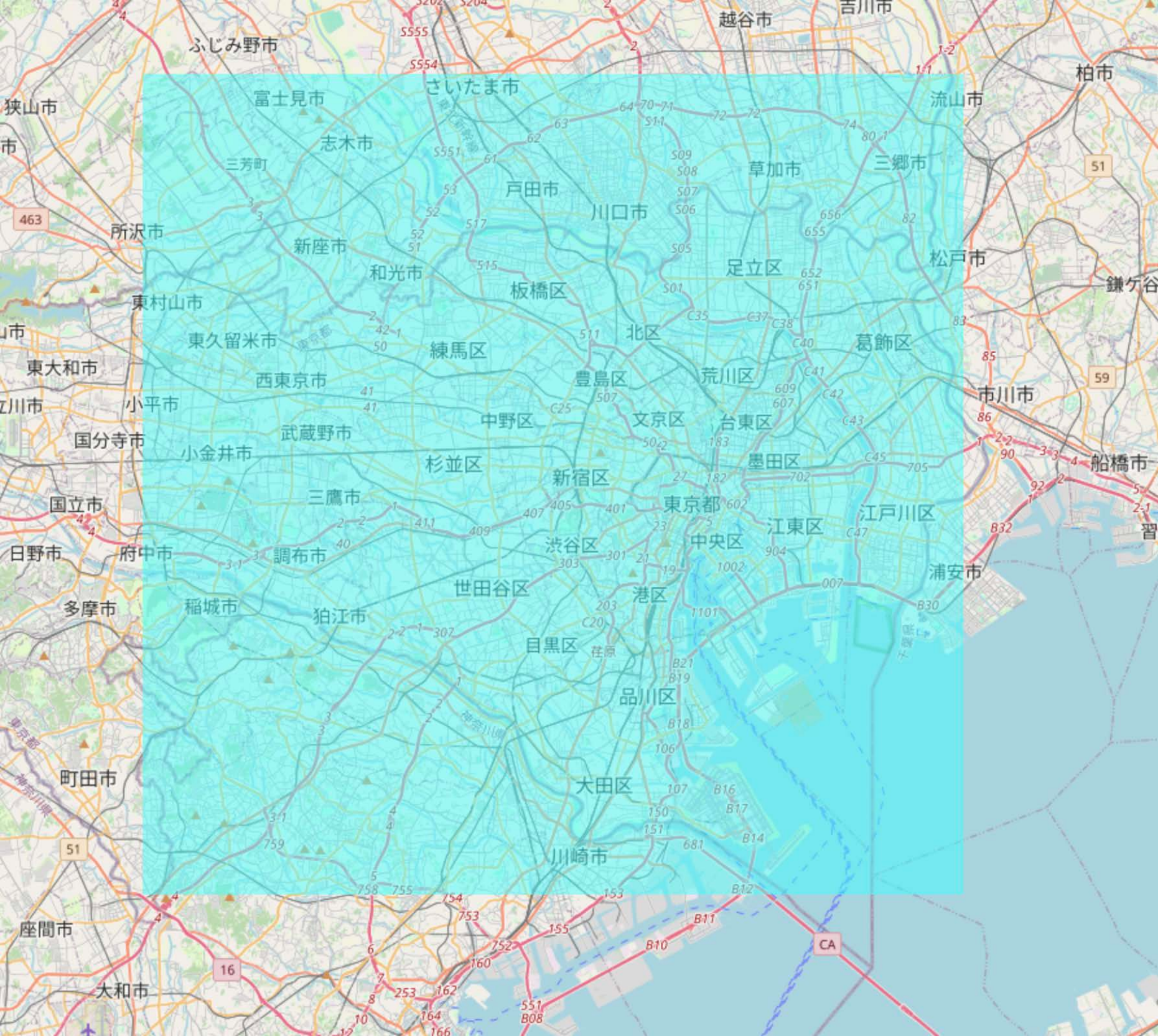}
        \caption{Tokyo, Japan (1540.18 $km^2$)}
        \label{fig:sub3}
    \end{subfigure}
    \hfill
    \begin{subfigure}[b]{0.48\linewidth}
        \includegraphics[width=\linewidth, height=4cm]{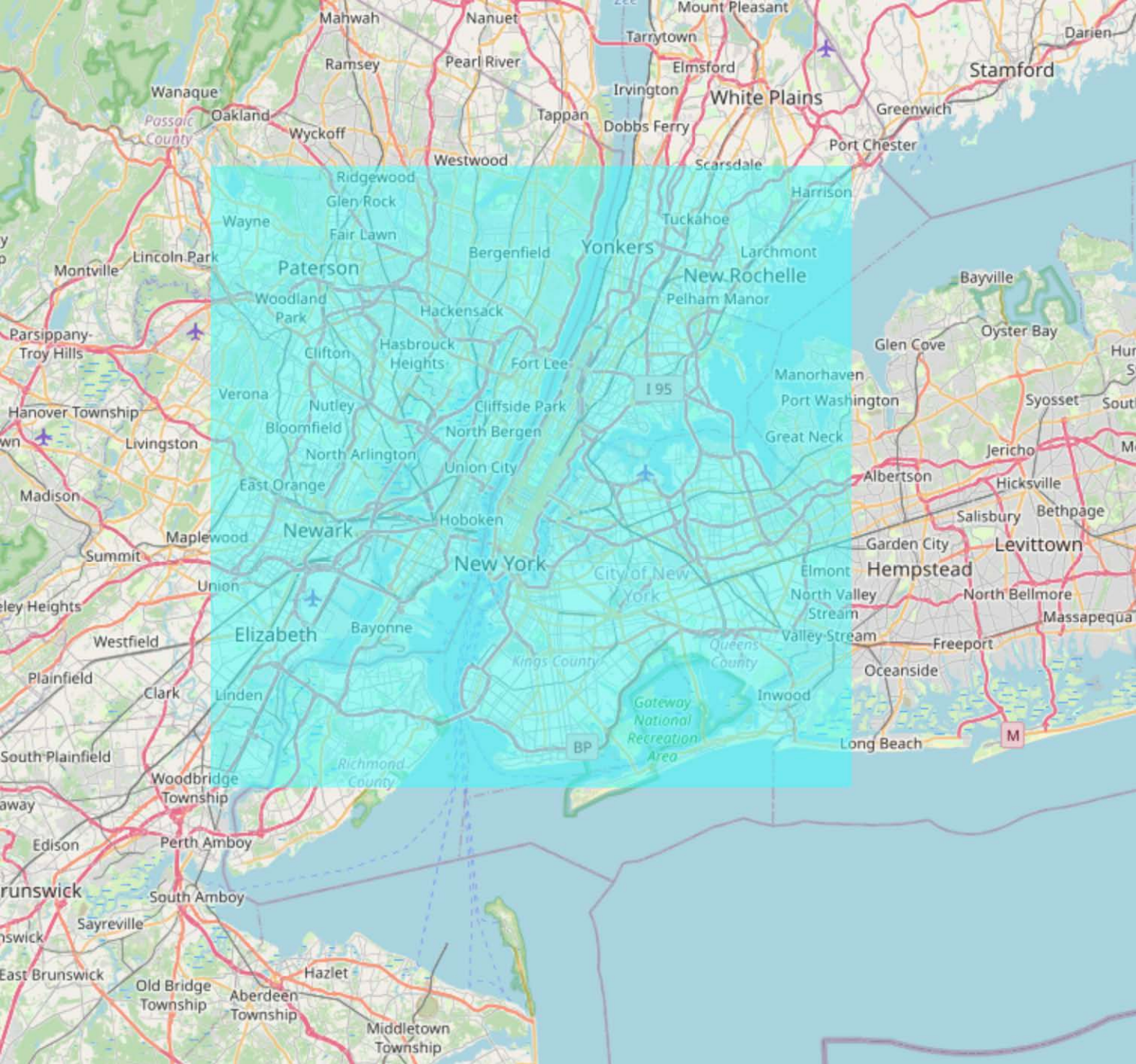}
        \caption{NYC, USA (2357.06 $km^2$)}
        \label{fig:sub4}
    \end{subfigure}

    \caption{The study areas (in light blue) of the datasets used in our paper.}
    \label{fig:study_area}
\end{figure}

We follow the same preprocessing protocol and similar experiment settings from \cite{chang2023contrastive} and keep trajectories in urban areas with the number of points ranging from 20 to 200. Table \ref{tab:data_summary}
provides the statistics of datasets after preprocessing. The study areas for all cities are shown in Fig. \ref{fig:study_area}. We select the same study area for T-Drive and GeoLife as they are both sampled in Beijing. We use 200,000 trajectories for Porto, 70,000 for T-Drive, 35000 for GeoLife, 2133 for TKY, and 513 for NYC as training sets. Each dataset has 10\% of data used for validation. Since there are large differences between the number of trajectories for each dataset, we select 100,000 trajectories for testing in Porto, 10,000 for T-Drive and GeoLife, 500 for TKY, and 147 for NYC. For downstream fine-tuning tasks that approximate the heuristic measures, we select 10,000 trajectories for Porto and T-Drive, and 5000 for GeoLife, where the selected trajectories are split by 7:1:2 for training, validation, and testing. We train our T-JEPA from scratch for Porto, T-Drive, and GeoLife datasets. To align the preprocessing with previous work \cite{chang2023contrastive}, most trajectories in TKY and NYC are filtered for being too short. Training on these two datasets from scratch will likely cause overfitted and biased models. Therefore, we load the pre-trained T-JEPA weights from Porto and continue the training on each of the TKY and NYC datasets. 

\noindent\textbf{Baselines.} Since this paper focuses on self-supervised learning methods, we compare our T-JEPA with other two baselines: t2vec \cite{li2018deep} and TrajCL \cite{chang2023contrastive}. t2vec pioneers self-supervised learning methods in trajectory similarity computation and TrajCL is the current state-of-the-art model on trajectory similarity computation in multiple datasets and experimental settings. We run these two models from their open-source code repositories with default parameters.

\begin{table}
\centering
\caption{Statistics of Datasets after preprocessing.}
\begin{tabular}{@{}llrr@{}}
\toprule
Data type & Dataset   & \#points   & \#trajectories \\ \midrule
\multirow{3}*{\makecell{GPS \\ trajectories}}  &  Porto     & 65,913,828 & 1,372,725    \\
  & T-Drive & 5,579,067 & 101,842 \\
  &  GeoLife   & 8,987,488 & 50,693           \\

  \hline
\multirow{2}*{\makecell{Check-in \\ sequences}} & Tokyo & 106,480 & 3048 \\
 & NYC & 28,858 & 734 \\\bottomrule
\end{tabular}
    \setlength{\abovedisplayskip}{3pt}
    \setlength{\belowcaptionskip}{-3 cm}
    \label{tab:data_summary}
\end{table}

% \begin{figure}[htp]
%     \centering

%     % First row of figures
%     \subfloat[Porto, Portugal (183.13 $km^2$)]{
%     \includegraphics[width=0.48\linewidth, height=4cm]{cities/porto.pdf}
%     \label{fig:sub1}
%     }
%     \hfill
%     \subfloat[Beijing, China (1949.26 $km^2$)]{
%     \includegraphics[width=0.48\linewidth, height=4cm]{cities/beijing.pdf}
%     \label{fig:sub2}
%     }
%     \vspace{0.1cm} % Space between rows
    
%     % Second row of figures
%     \subfloat[Tokyo, Japan (1540.18 $km^2$)]{
%     \includegraphics[width=0.48\linewidth, height=4cm]{cities/tokyo.pdf}
%     \label{fig:sub3}
%     }
%     \hfill
%     \subfloat[New York City, USA (2357.06 $km^2$)]{
%     \includegraphics[width=0.48\linewidth, height=4cm]{cities/newyork.pdf}
%     \label{fig:sub4}
%     }

% \caption{The study areas (in light blue) of the datasets used in our paper. }
% \label{fig:study_area}
% \end{figure}

\noindent\textbf{Implementation details.} We use Adam Optimizer for training and optimizing the model parameters, except for the target encoder which updates via the exponential moving average of the parameters of the context encoder. The maximum number of training epochs is 20 and the learning rate is 0.0001 which decays by half every 5 epochs. The early stopping is set if there are no improvements with training loss in 5 successive epochs. The embedding dimension $d$ is 256 and the batch size is 64. We use 3-layer Transformer Encoders for both context and target encoders with the number of attention heads set to 4. We use a 2-layer Transformer Decoder for the predictor with the number of attention heads set to 8. Positional encoding for both encoders and the decoder is learnable. We set the resampling masking ratio to be selected from \{10\%, 20\%, 30\%\} and the number of sampled targets $M$ to 4 for each trajectory. The successive sampling probability $p$ is set to 50\% and the initial context sampling ratio $p_{\gamma}$ is set to range from 85\% to 100\%. All experiments are conducted on servers with Nvidia V100 GPUs.

\subsection{Quantitative Evaluation}

\subsubsection{Most similar trajectory search} \label{sec:traj_search}
We test the capability of T-JEPA to find the most similar trajectory by following the same settings of previous works \cite{li2018deep,liu2022cstrm,chang2023contrastive}. We construct a Query trajectory set $Q$ and a database trajectory $D$ for the testing set. $Q$ has 1,000 trajectories for Porto, T-Drive, and GeoLife, 50 for TKY, and 20 for NYC. And $D$ has 100,000 trajectories for Porto, 10,000 for T-Drive and Geolife, 500 for TKY, and 147 for NYC. 

For each query trajectory $q\in Q$, we create two sub-trajectories containing the odd-indexed points and even-indexed points of $q$. We denote these two trajectories $q_{a}=\{X_{1}, X_{3}, X_{5}, \ldots\}$ and $q_{b}=\{X_{2}, X_{4}, X_{6}, \ldots\}$. We separate them by putting $q_{a}$ into the query set $Q$ and putting $q_{b}$ into the database $D$. Other trajectories in $D$ will be randomly filled from the testing set. The intuition of the setting is that two variants from the same trajectory can reflect overall similar moving patterns in shape, length, and sampling rates. We use T-JEPA to encode the query and database trajectories to compare them on their representations and calculated similarities will be sorted in descending order. We report the mean rank of each $q_{b}$ found by $q_{a}$, the most accurate match should have the first rank.

Since the database size varies for each dataset, we choose \{20\%, 40\%, 60\%, 80\%, 100\%\} of the total database size. Table \ref{tab:db_size} shows the \textbf{mean rank} from different models under such varying database size, where the down arrows in the first row mean the lower the values the better the results. Our T-JEPA outperforms the baselines in 4 of 5 datasets, producing more consistent low mean ranks close to 1. Although TrajCL achieves slightly better mean ranks in Porto, the overall mean rank is only 0.044 lower than T-JEPA. This shows our T-JEPA using $AdjFuse$ under the JEPA framework is competitive with TrajCL using Dual-feature attention under the contrastive learning framework. Besides, we outperform TrajCL on T-drive by 0.114 and on GeoLife by 0.158. Moreover, for Foursquare data with much sparser check-in trajectories, we outperform t2vec and TrajCL each by 6.88 times and 2.92 times in TKY, and 8.05 times and 2.51 times in NYC. Proving that T-JEPA can be more generalized and robust to low and irregular sampled mobility patterns. The constant mean ranks generated by T-JEPA in NYC are probably due to the small size of the database (147), reflecting the stable search for the most similar trajectories. Compared with Porto, the other 4 datasets have much larger regions with sparser trajectories, with the trajectory point sampling intervals being longer for T-Drive, TKY, and NYC. This demonstrates our T-JEPA can provide larger receptive fields with richer information for each trajectory point by the $AdjFuse$ module, with a better trajectory semantic understanding of sparse trajectory point correlations.

\begin{table}
\centering
\small
\caption{Mean rank of the ground truths in varying DB sizes}
\begin{tabular}{l@{\hspace{0.1em}} | l@{\hspace{0.1em}} | c@{\hspace{0.3em}}  | c@{\hspace{0.3em}}  | c@{\hspace{0.3em}}  | c@{\hspace{0.3em}} | c@{\hspace{0.3em}} }
 \hline
 Dataset & 
 Method &
 20\%$\downarrow$ &
 40\%$\downarrow$ &
 60\%$\downarrow$ &
 80\%$\downarrow$ &
 100\%$\downarrow$ \\
 \hline\hline
 \multirow{3}*{Porto} & t2vec & 2.190 & 3.296 & 4.508 & 5.989 & 7.163 \\
 % & CL-TSim & 2.750 & 5.737 & 8.852 & 12.095 & 15.395 \\
 & TrajCL & \textbf{1.004} & \textbf{1.007} & \textbf{1.008} & \textbf{1.011} & \textbf{1.014} \\
 & T-JEPA & 1.029 & 1.048 & 1.053 & 1.061 & 1.074 \\
\hline
\multirow{3}*{T-Drive} & t2vec & 3.377 & 3.746 & 4.055 & 4.385 & 4.806 \\
 & TrajCL & 1.111 & 1.128 & 1.146 & 1.177 & 1.201 \\
 & T-JEPA & \textbf{1.032} & \textbf{1.034} & \textbf{1.036} & \textbf{1.045} & \textbf{1.049} \\
 \hline
\multirow{3}*{GeoLife} & t2vec & 4.580 & 4.634 & 4.709 & 4.815 & 4.922 \\
 & TrajCL & 1.130 & 1.168 & 1.195 & 1.234 & 1.256 \\
 & T-JEPA & \textbf{1.019} & \textbf{1.034} & \textbf{1.036} & \textbf{1.040} & \textbf{1.047} \\
\hline
\multirow{3}*{Tokyo} & t2vec & 5.380 & 7.980 & 10.300 & 13.240 & 16.340 \\
& TrajCL & 2.420 & 3.620 & 4.580 & 5.480 & 6.460 \\
& T-JEPA & \textbf{1.120} & \textbf{1.320} & \textbf{1.560} & \textbf{1.760} & \textbf{1.980} \\
\hline
\multirow{3}*{NYC} & t2vec & 5.600 & 9.650 & 10.500 & 12.450 & 15.300 \\
& TrajCL & 2.300 & 2.800 & 3.200 & 3.900 & 4.500 \\
& T-JEPA & \textbf{1.250} & \textbf{1.350} & \textbf{1.350} & \textbf{1.350} & \textbf{1.350} \\
\hline
\end{tabular} 
\label{tab:db_size}
\end{table}

\subsubsection{Robustness}
We test our model with down-sampled and distorted trajectories separately to evaluate the robustness towards low and irregular sampled data, as well as noisy data. For down-sampling, We randomly mask points in both $Q$ and $D$ with a probability $\rho_{s}$ ranging from 0.1 to 0.5. For distortion, we shift the point coordinates to create noisy data, where the number of points distorted is controlled by another probability $\rho_{d}$. The database size for experiments with both down-sampling and distortion settings is the same as the total size in Sec. \ref{sec:traj_search} for each dataset. 

Table \ref{tab:downsampling} shows the comparisons for various down-sampling rates. We can find that TrajCL still outperforms T-JEPA in Porto, but experiences a slump when increasing the down-sampling rate $\rho_{s}$ from 0.4 to 0.5, while we produce a lower mean rank in this case. For T-Drive, we outperform TrajCL from the down-sampling rate $\rho_{s}$ of 0.1 to 0.3 but produce higher mean ranks for the rates of 0.4 and 0.5. This is because one of the data augmentation schemes of TrajCL is trajectory down-sampling, resulting in better performances in such cases. Besides, we achieve the best performance in the other 3 datasets, demonstrating our competitive performance on down-sampled most similar trajectory searches without manually creating such training cases.

Table \ref{tab:distortion} shows the comparison for various distortion rates. Except for Porto where our overall mean rank is 0.073 higher than TrajCL, we outperform TrajCL and t2vec for all other 4 datasets. Our T-JEPA outperforms TrajCL by 1.62 times in T-Drive, 10.92 times in GeoLife, 3.04 times in TKY, and 2.53 times in NYC. As we can notice, the mean ranks may not increase with higher distortion rates, which can be due to the random distortions over all trajectories. As explained by \cite{chang2023contrastive}, the actual relative similarities might change for each query trajectory and its ground truths. The overall best performance in varying distortion rates exhibits the robustness of T-JEPA over noisy points by its unique predictive module which emphasizes the high-level semantic understanding in representation space.

\begin{table}
\centering
\small
\caption{Mean rank versus the down-sampling rates}
\begin{tabular}{l@{\hspace{0.1em}} | l@{\hspace{0.1em}} | c@{\hspace{0.3em}}  | c@{\hspace{0.3em}}  | c@{\hspace{0.3em}}  | c@{\hspace{0.3em}} | c@{\hspace{0.3em}} }
 \hline
 Dataset & 
 Method &
 0.1$\downarrow$&
 0.2$\downarrow$&
 0.3$\downarrow$&
 0.4$\downarrow$&
 0.5$\downarrow$\\
 \hline\hline
 \multirow{3}*{Porto} & t2vec & 7.305 & 7.498 & 15.651 & 21.322 & 69.817 \\
 & TrajCL & \textbf{1.038} & \textbf{1.333} & \textbf{1.856} & \textbf{5.654} & 83.777 \\
 & T-JEPA & 1.514 & 3.840 & 11.128 & 28.362 & \textbf{61.422} \\
\hline
\multirow{3}*{T-Drive} & t2vec & 5.389 & 5.428 & 4.383 & 10.398 & 13.191 \\
 & TrajCL & 1.198 & 1.492 & 2.544 & \textbf{4.680} & \textbf{12.647} \\
 & T-JEPA & \textbf{1.088} & \textbf{1.211} & \textbf{1.481} & 5.040 & 19.110 \\
 \hline
\multirow{3}*{GeoLife} & t2vec & 5.198 & 5.042 & 5.290 & 5.948 & 5.664 \\
 & TrajCL & 1.224 & 1.336 & 1.441 & \textbf{1.790} & 2.197 \\
 & T-JEPA & \textbf{1.057} & \textbf{1.182} & \textbf{1.179} & 1.826 & \textbf{1.425} \\
\hline
\multirow{3}*{Tokyo} & t2vec & 10.040 & 12.240 & 24.420 & 17.980 & 39.900 \\
& TrajCL & 5.760 & 3.260 & 25.780 & 3.260 & 16.960 \\
& T-JEPA & \textbf{2.920} & \textbf{2.380} & \textbf{2.560} & \textbf{2.760} & \textbf{4.120} \\
\hline
\multirow{3}*{NYC} & t2vec & 14.100 & 14.500 & 18.700 & 17.650 & 18.050 \\
& TrajCL & 1.550 & 6.300 & 2.200 & \textbf{1.700} & 6.700 \\
& T-JEPA & \textbf{1.250} & \textbf{1.600} & \textbf{1.150} & 1.750 & \textbf{2.050} \\
\hline
\end{tabular}
\label{tab:downsampling}
\end{table}

\begin{table}
\centering
\small
\caption{Mean rank versus the distortion rates}
\begin{tabular}{l@{\hspace{0.1em}} | l@{\hspace{0.1em}} | c@{\hspace{0.3em}}  | c@{\hspace{0.3em}}  | c@{\hspace{0.3em}}  | c@{\hspace{0.3em}} | c@{\hspace{0.3em}} }
 \hline
 Dataset & 
 Method &
 0.1$\downarrow$ &
 0.2$\downarrow$ & 
 0.3$\downarrow$ &
 0.4$\downarrow$ &
 0.5$\downarrow$ \\
 \hline\hline
 \multirow{3}*{Porto} & t2vec & 7.134 & 9.774 & 7.888 & 6.891 & 6.953 \\
 & TrajCL & \textbf{1.017} & \textbf{1.029} & \textbf{1.036} & \textbf{1.060} & \textbf{1.022} \\
 & T-JEPA & 1.097 & 1.084 & 1.115 & 1.110 & 1.123 \\
 \hline
 \multirow{3}*{T-Drive} & t2vec & 4.719 & 4.601 & 4.491 & 4.588 & 4.510 \\
 & TrajCL & 1.267 & 3.320 & 1.355 & 1.513 & 1.179 \\
 & T-JEPA & \textbf{1.054} & \textbf{1.061} & \textbf{1.069} & \textbf{1.067} & \textbf{1.078} \\
 \hline
 \multirow{3}*{GeoLife} & t2vec & 5.628 & 5.670 & 5.412 & 5.306 & 5.838 \\
 & TrajCL & 7.973 & 19.266 & 12.397 & 10.560 & 11.035 \\
 & T-JEPA & \textbf{1.047} & \textbf{1.093} & \textbf{1.101} & \textbf{1.154} & \textbf{1.197} \\
\hline
\multirow{3}*{Tokyo} & t2vec & 14.240 & 14.780 & 19.300 & 15.900 & 17.640 \\
& TrajCL & 5.860 & 7.540 & 7.260 & 4.380 & 5.520 \\
& T-JEPA & \textbf{2.100} & \textbf{1.980} & \textbf{1.920} & \textbf{2.100} & \textbf{1.960} \\
\hline
\multirow{3}*{NYC} & t2vec & 15.350 & 15.600 & 15.900 & 15.500 & 15.650 \\
& TrajCL & 4.950 & 2.250 & 1.650 & 5.150 & 2.950 \\
& T-JEPA & \textbf{1.350} & \textbf{1.300} & \textbf{1.300} & \textbf{1.350} & \textbf{1.400} \\
\hline
\end{tabular}
\label{tab:distortion}
\end{table}

\subsubsection{Approximating heuristic measures}
Approximating heuristic measures is a recently proposed experiment by TrajCL \cite{chang2023contrastive} which can reflect the generalization ability of the learned representations. Specifically, after we obtain the trained encoder for T-JEPA and other baselines, we concatenate a 2-layer MLP decoder with the same embedding dimension $d$. Compared to the previous setting from TrajCL that fine-tunes the decoder and the last layer of the encoder, we freeze the entire encoder and only train the decoder. The reason why we use this setting is that it contributes to a clearer generalization evaluation by ensuring consistent feature extraction from the encoder. By only tuning the decoder parameters, we can determine if the learned representations are powerful enough to capture the high-level semantic understanding from the pre-trained encoder.

\begin{table*}
\centering
\small
\caption{Comparisons with fine-tuning 2-layer MLP decoder.}
\begin{tabular}{l@{\hspace{0.1em}} | l@{\hspace{0.1em}} | c@{\hspace{0.3em}}  c@{\hspace{0.3em}}  c@{\hspace{0.3em}} | c@{\hspace{0.3em}}  c@{\hspace{0.3em}}  c@{\hspace{0.3em}} | c@{\hspace{0.3em}}  c@{\hspace{0.3em}}  c@{\hspace{0.3em}} |   c@{\hspace{0.3em}}  c@{\hspace{0.3em}} c@{\hspace{0.3em}}| c@{\hspace{0.2em}}}
 \hline
 \multirow{2}*{Dataset} & 
 % \multirow{2}*{Category} & 
 \multirow{2}*{Method} &
 \multicolumn{3}{c|}{EDR} & 
 \multicolumn{3}{c|}{LCSS} &
 \multicolumn{3}{c|}{Hausdorff} &
 \multicolumn{3}{c|}{Fréchet} &
 \multirow{2}*{Average} \\
 \cline{3-14} & & HR@5$\uparrow$ & HR@20$\uparrow$ & R5@20$\uparrow$ & HR@5$\uparrow$ & HR@20$\uparrow$ & R5@20$\uparrow$ & HR@5$\uparrow$ & HR@20$\uparrow$ & R5@20$\uparrow$ & HR@5$\uparrow$ & HR@20$\uparrow$ & R5@20$\uparrow$ &  \\
 \hline\hline
\multirow{3}*{Porto} & t2vec & 0.002 & 0.012 & 0.011 & 0.004 & 0.010 & 0.012 & 0.003 & 0.010 & 0.010 & 0.004 & 0.011 & 0.011 & 0.008 \\
& TrajCL & 0.137 & 0.179 & 0.301 & 0.329 & 0.508 & 0.663 & 0.456 & 0.574 & 0.803 & 0.412 & 0.526 & 0.734 & 0.468 \\
& T-JEPA & \textbf{0.154} & \textbf{0.194} & \textbf{0.336} & \textbf{0.365} & \textbf{0.551} & \textbf{0.713} & \textbf{0.525} & \textbf{0.633} & \textbf{0.869} & \textbf{0.433} & \textbf{0.565} & \textbf{0.771} & \textbf{0.509} \\
\hline
\multirow{3}*{T-Drive} & t2vec & 0.007 & 0.015 & 0.020 & 0.008 & 0.016 & 0.019 & 0.006 & 0.013 & 0.017 & 0.006 & 0.013 & 0.016 & 0.013 \\
& TrajCL & \textbf{0.094} & 0.131 & 0.191 & 0.159 & 0.289 & 0.366 & \textbf{0.173} & \textbf{0.256} & \textbf{0.356} & \textbf{0.138} & \textbf{0.187} & \textbf{0.274} &  0.218 \\
& T-JEPA & \textbf{0.094} & \textbf{0.147} & \textbf{0.215} & \textbf{0.205} & \textbf{0.366} & \textbf{0.469} & 0.158 & 0.229 & 0.329 & 0.125 & 0.159 & 0.249 & \textbf{0.229} \\
\hline
\multirow{3}*{GeoLife} & t2vec & 0.008 & 0.024 & 0.023 & 0.011 & 0.025 & 0.030 & 0.011 & 0.030 & 0.033 & 0.013 & 0.029 & 0.035 &  0.023 \\
& TrajCL & 0.193 & 0.363 & 0.512 & 0.232 & 0.484 & 0.584 & 0.479 & 0.536 & 0.745 & 0.398 & \textbf{0.463} & 0.708 & 0.475 \\
& T-JEPA & \textbf{0.195} & \textbf{0.383} & \textbf{0.527} & \textbf{0.242} & \textbf{0.515} & \textbf{0.586} & \textbf{0.606} & \textbf{0.656} & \textbf{0.857} & \textbf{0.488} & 0.406 & \textbf{0.731} & \textbf{0.516} \\
\hline
\end{tabular}
\label{tab:finetune}
\end{table*}

We report hit ratios \textbf{HR@5} and \textbf{HR@20} representing the correct matches between top-5 ground truths and top-5 predictions, and the recall \textbf{R5@20} of the correct matching in top-5 trajectories from predicted 20 trajectories. We test the approximation performance on EDR, LCSS, Hausdorff, and Discrete Fréchet. We also calculate an \textbf{average} value of all reported results for each model under all 4 heuristic measures. Since there is too little data left in TKY and NYC datasets, we only fine-tune the models in Porto, T-Drive, and GeoLife. 

\begin{figure*}[htbp]
    \centering
    \subfloat[TrajCL Visualizations]{
    \label{subfig:trajcl1}
    \includegraphics[width=3cm,height=2cm]{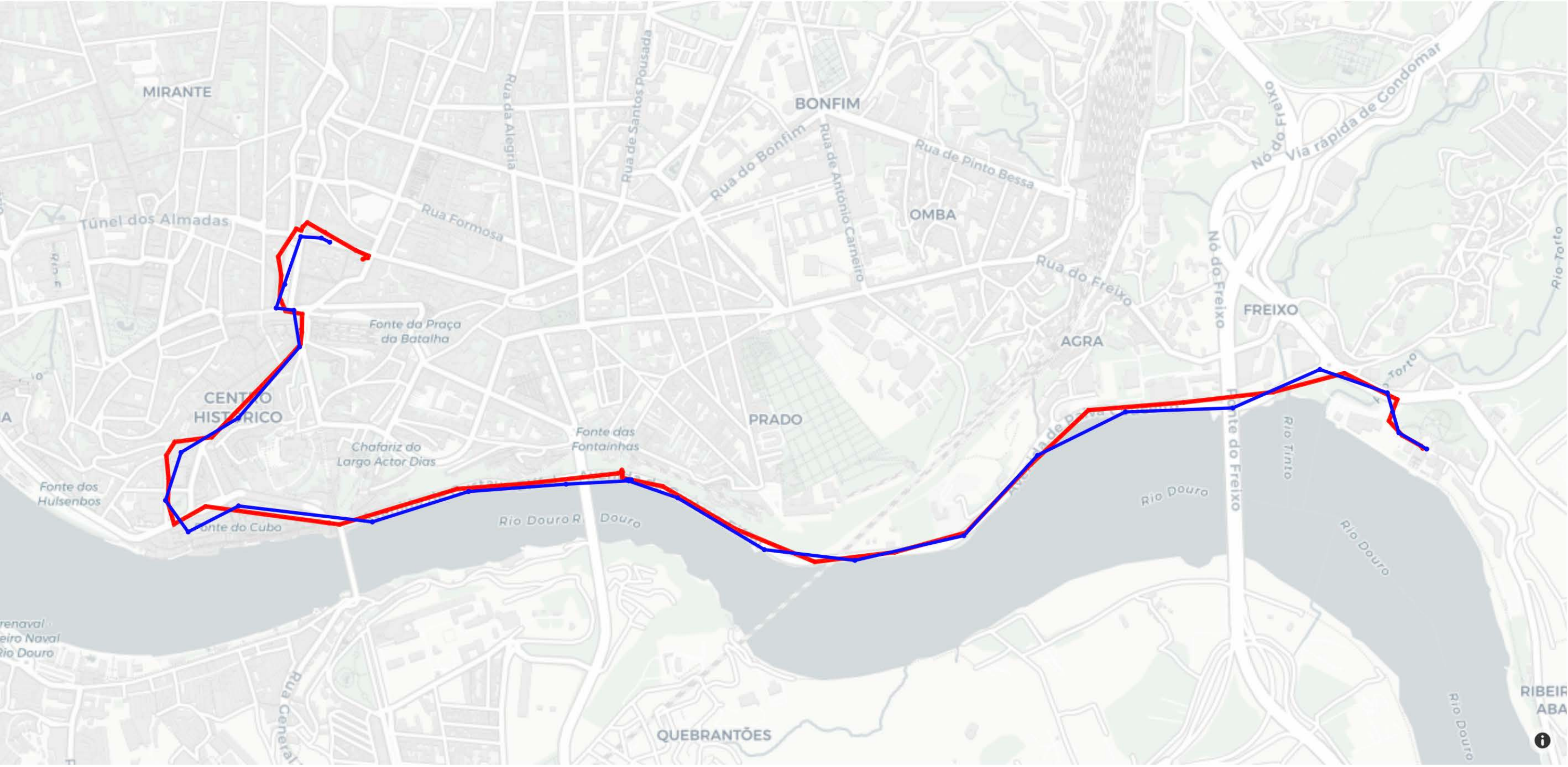}
    \includegraphics[width=3cm,height=2cm]{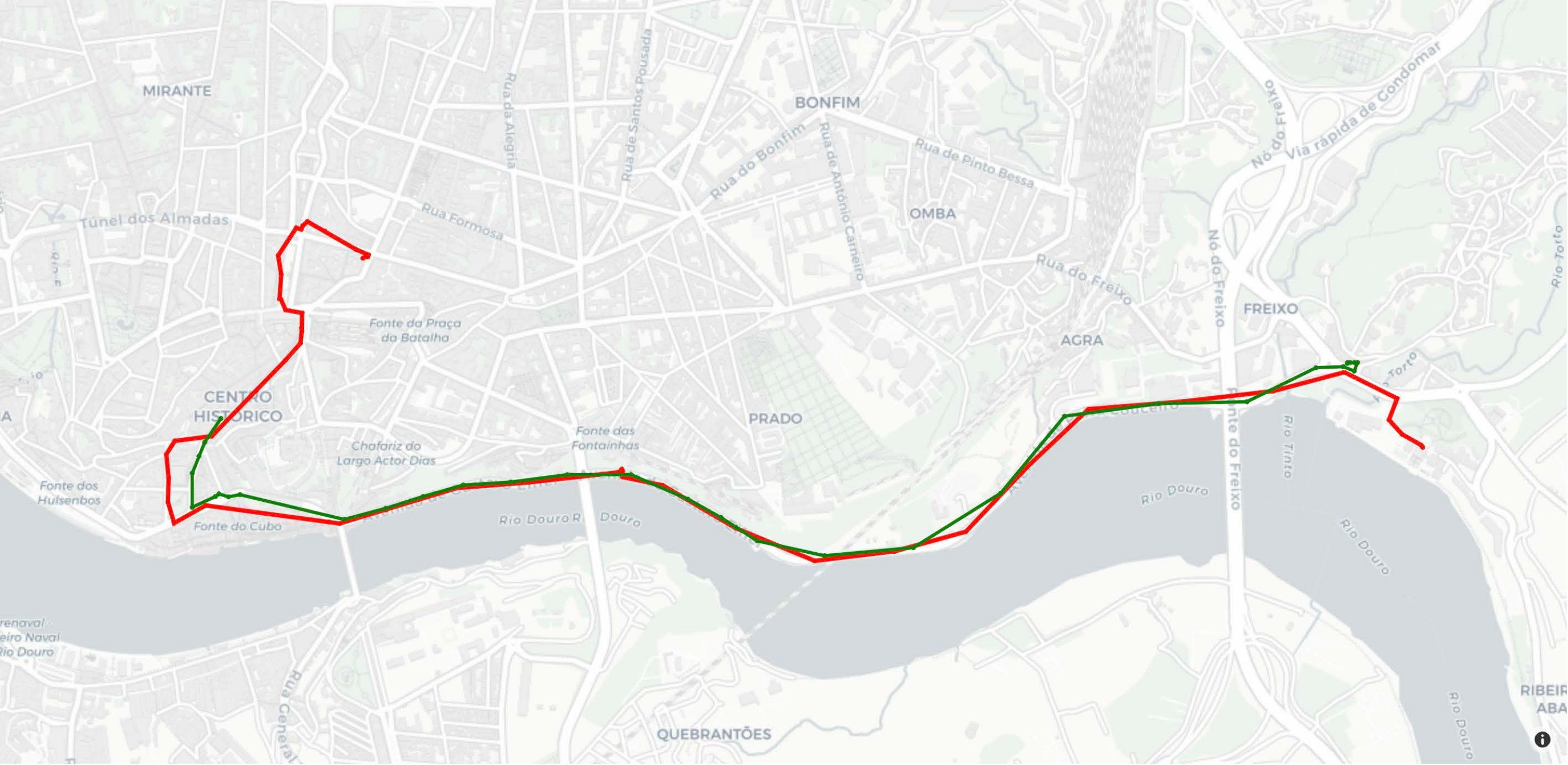}
    \includegraphics[width=3cm,height=2cm]{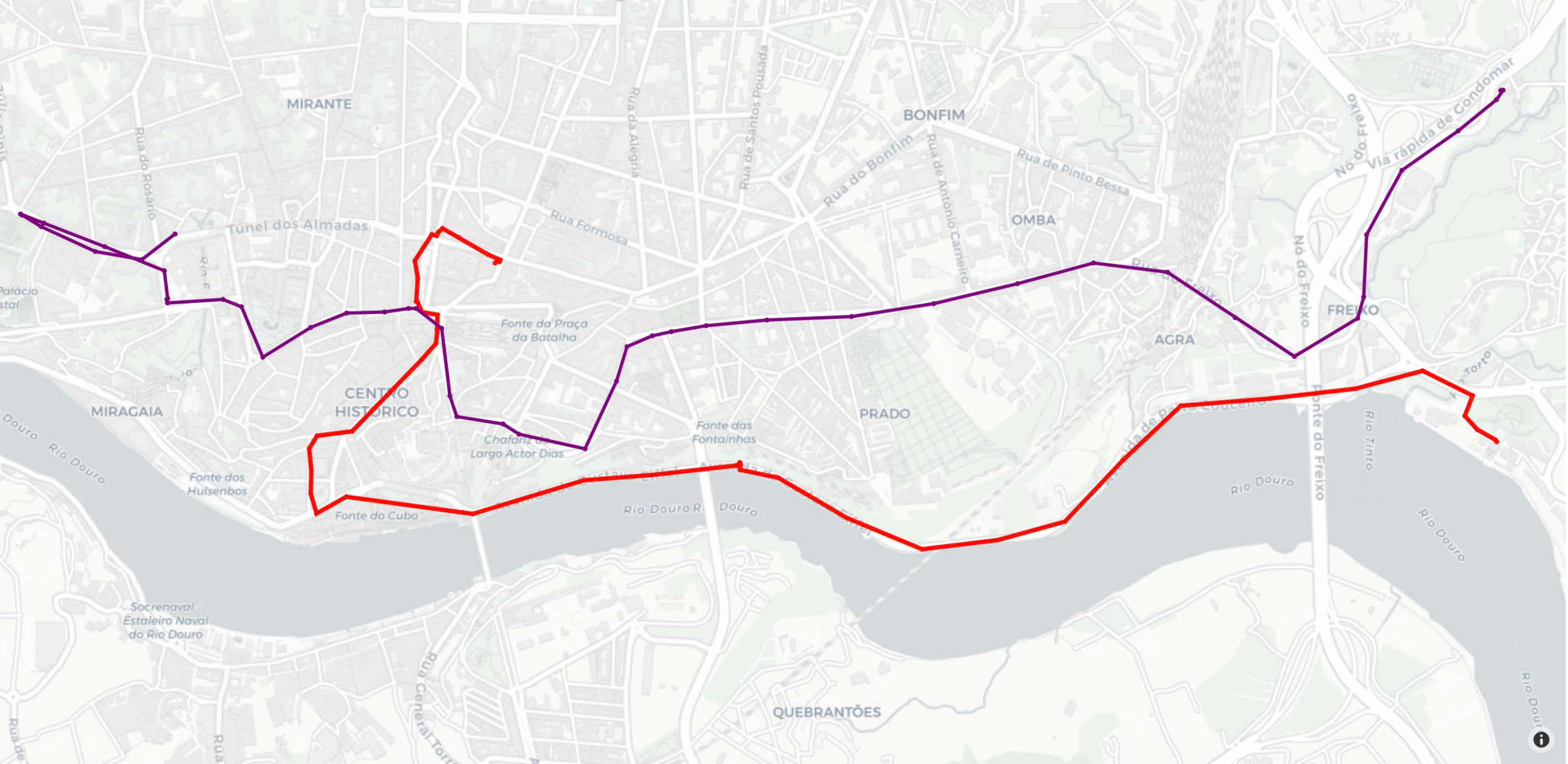}
    \includegraphics[width=3cm,height=2cm]{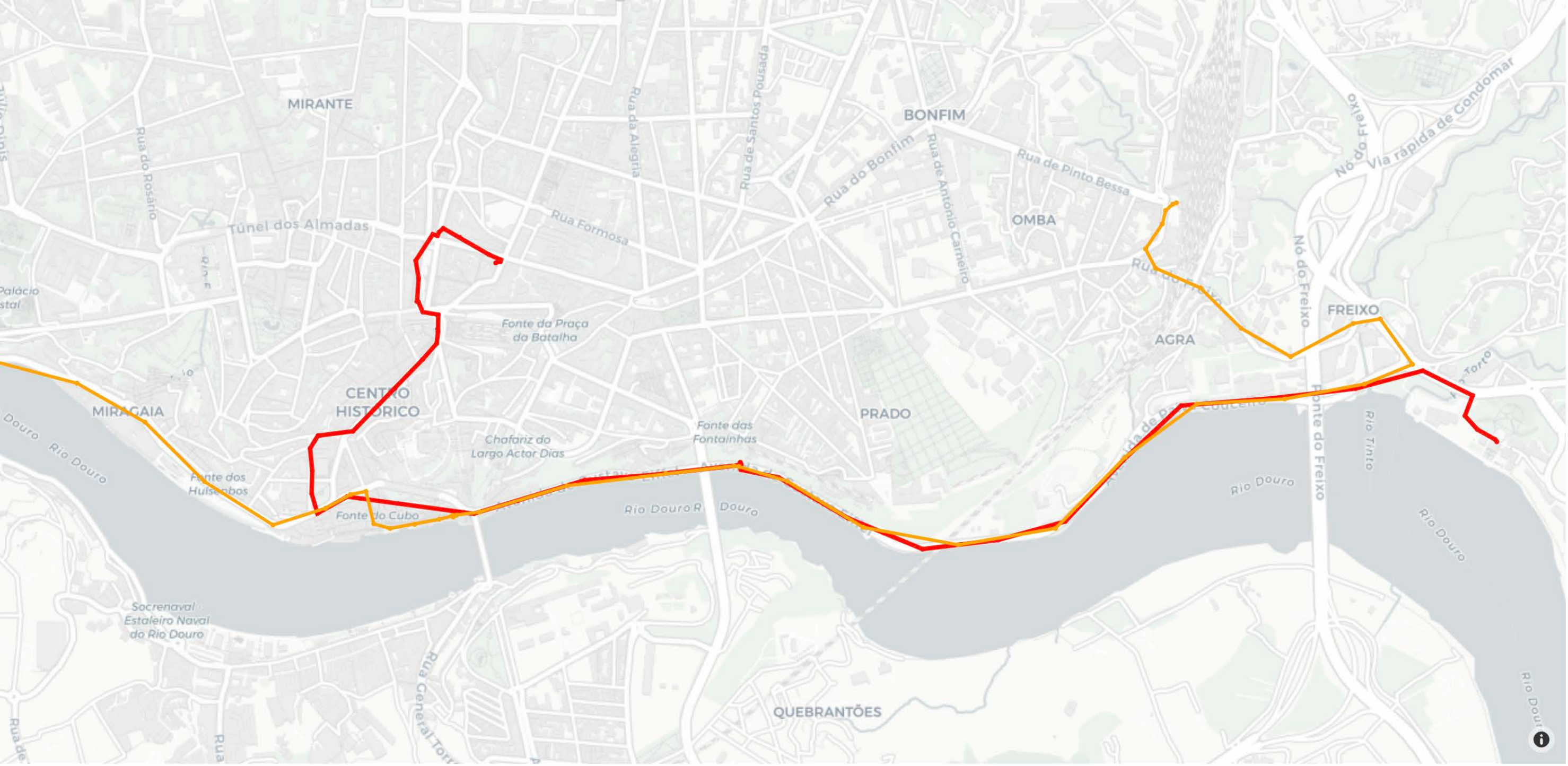}
    \includegraphics[width=3cm,height=2cm]{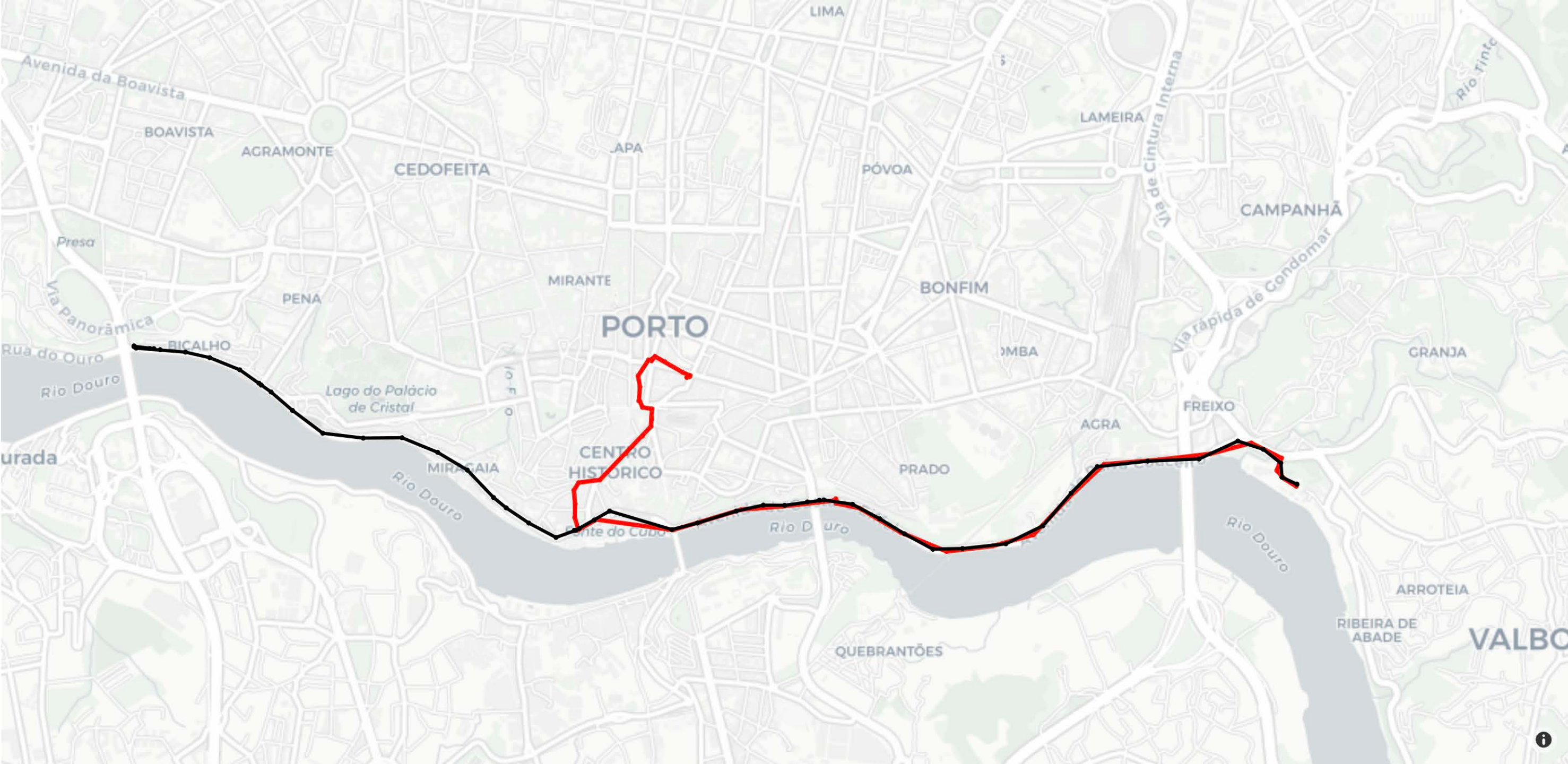}
    \includegraphics[width=2cm,height=1.5cm]{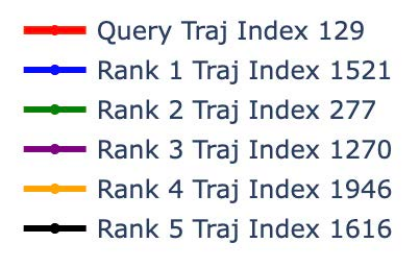}
    }
    
    \subfloat[T-JEPA Visualizations]{
    \label{subfig:tjepa1}
    \includegraphics[width=3cm,height=2cm]{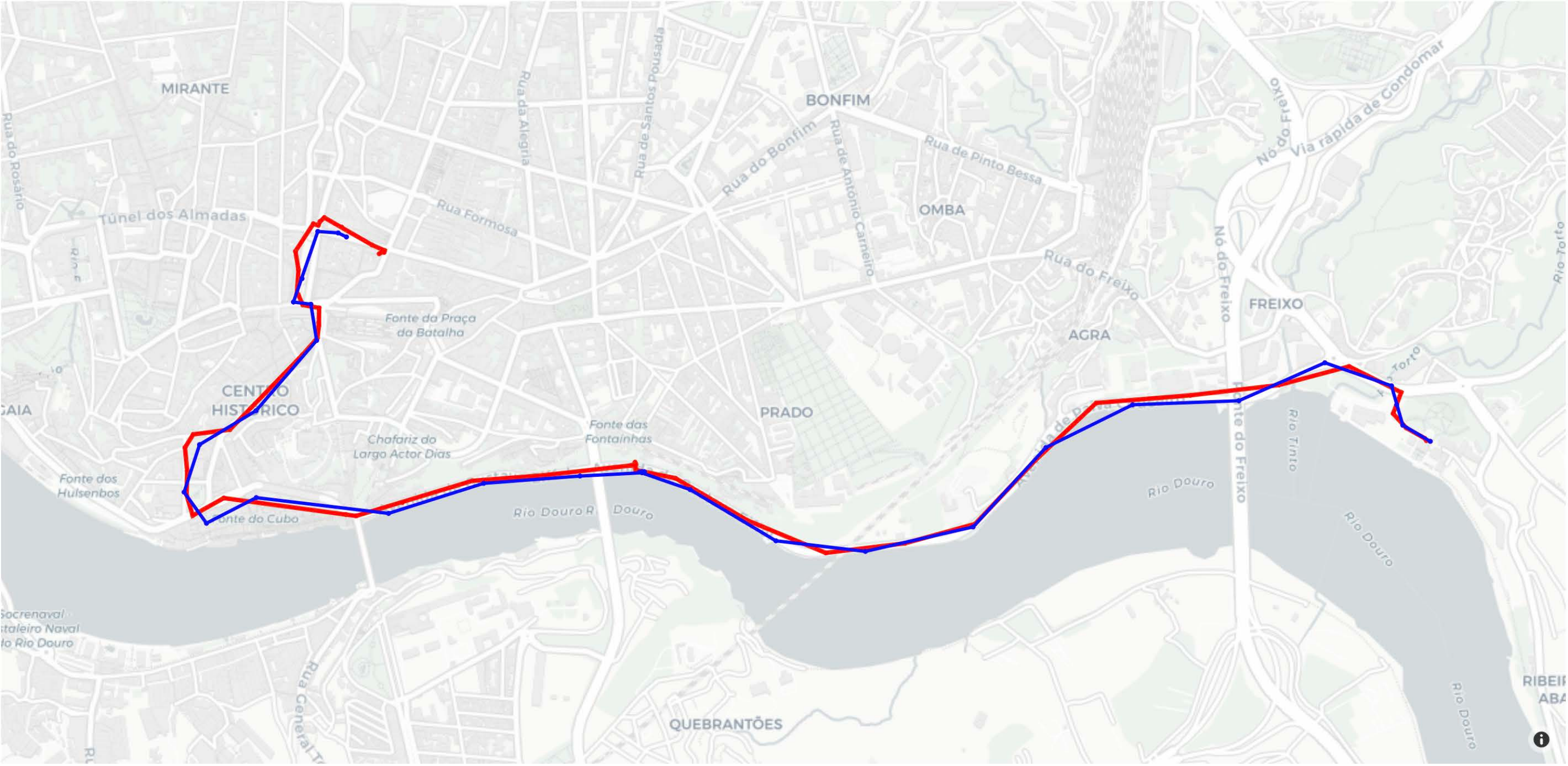}
    \includegraphics[width=3cm,height=2cm]{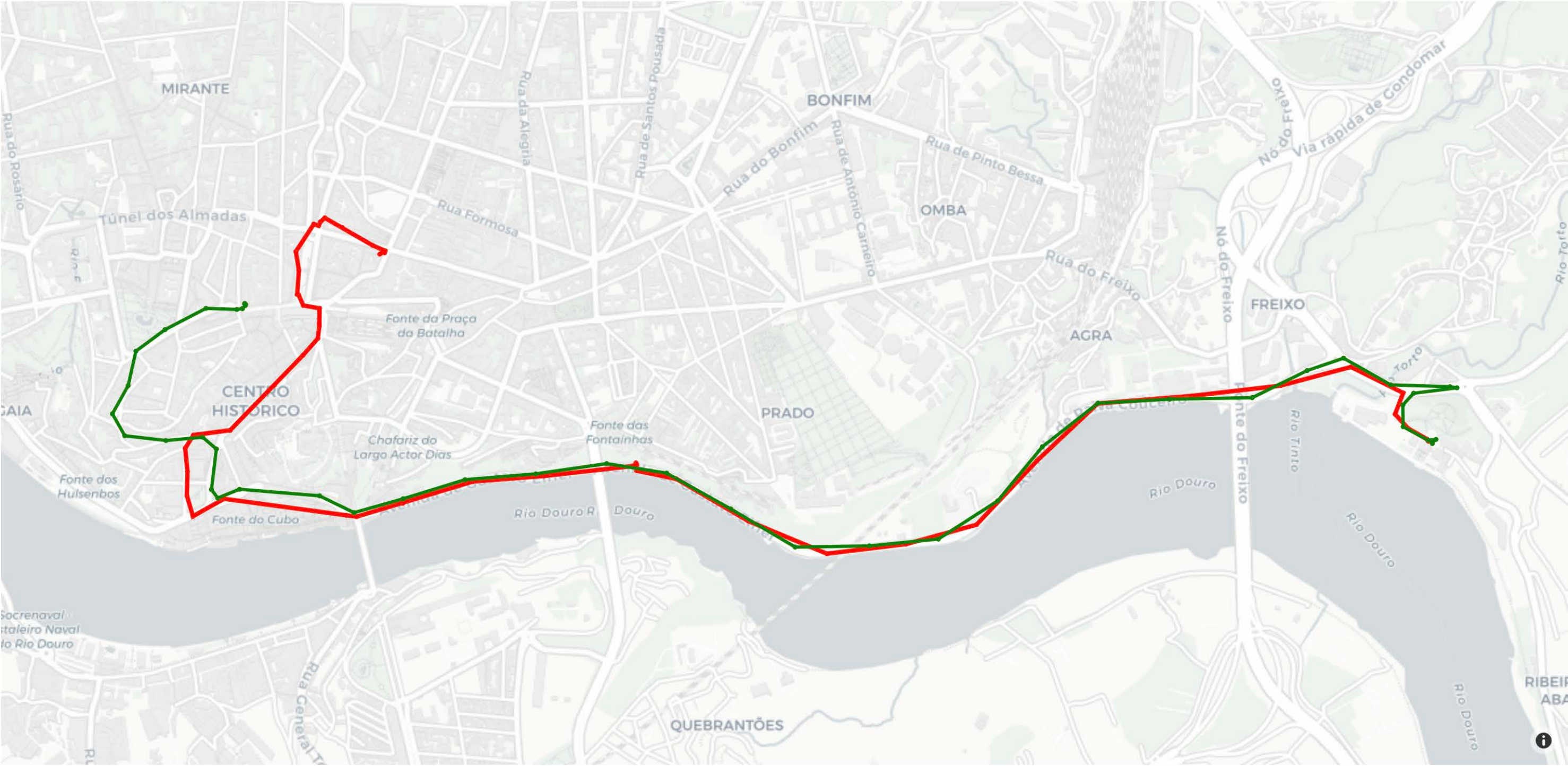}
    \includegraphics[width=3cm,height=2cm]{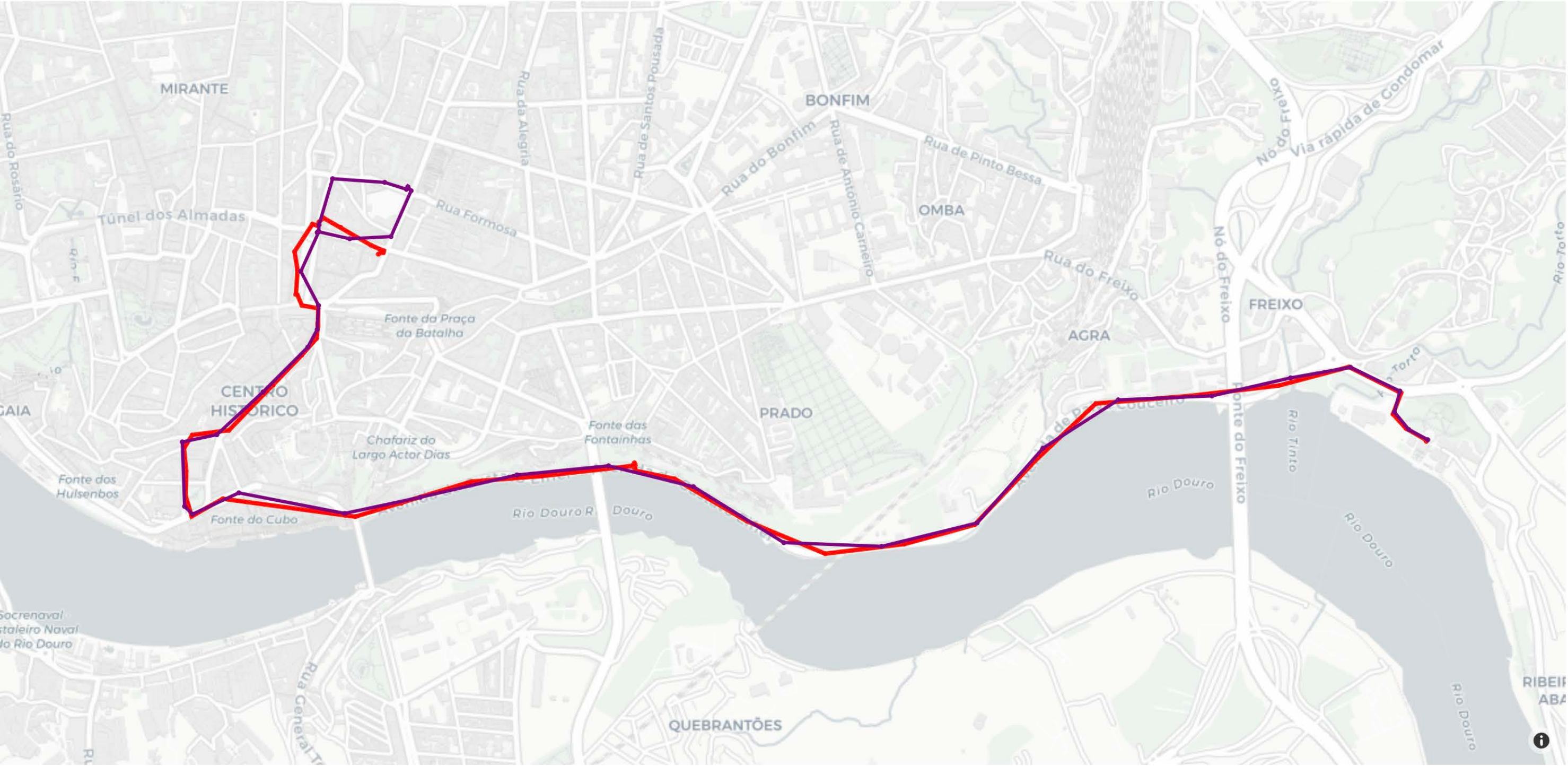}
    \includegraphics[width=3cm,height=2cm]{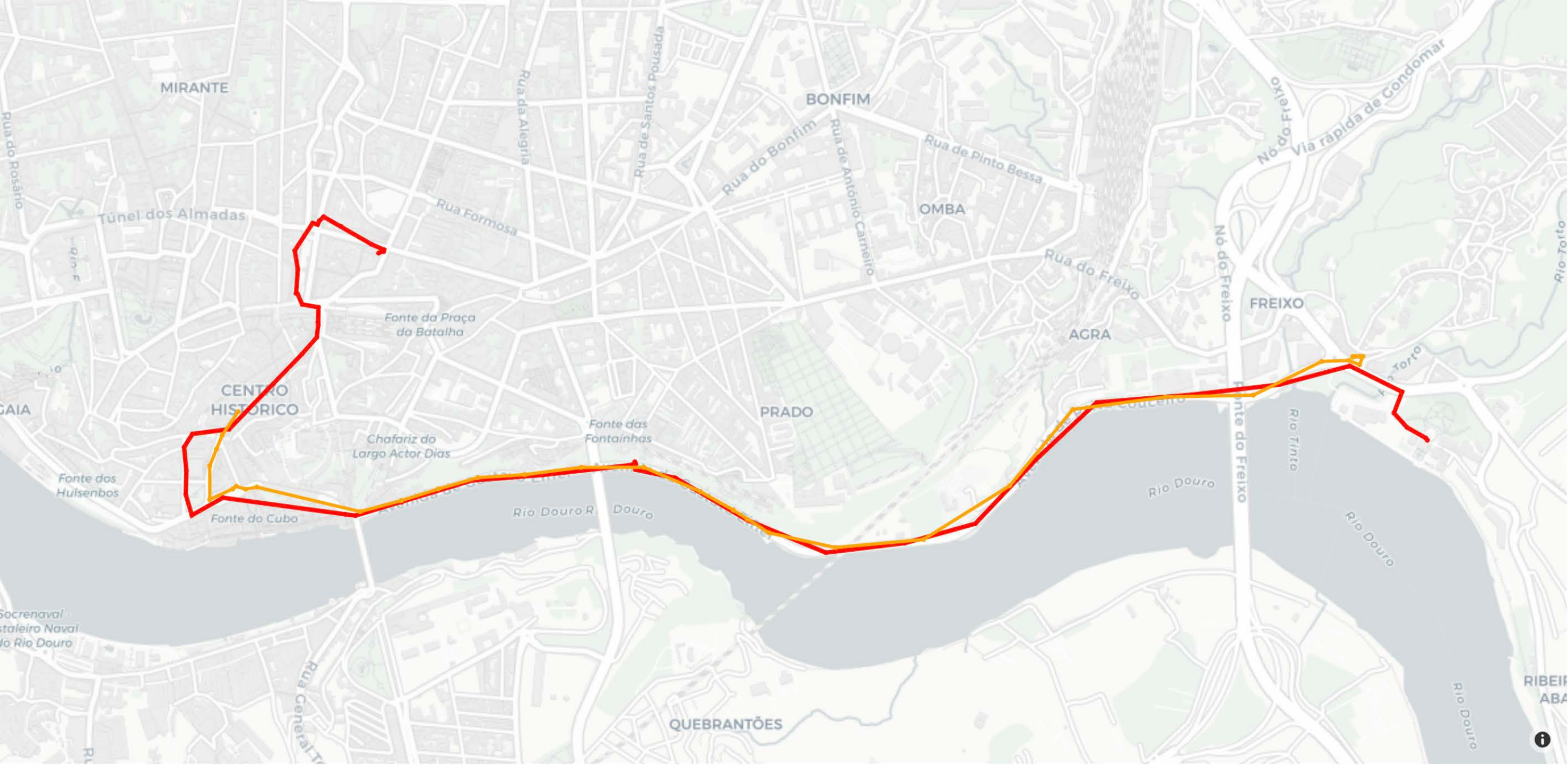}
    \includegraphics[width=3cm,height=2cm]{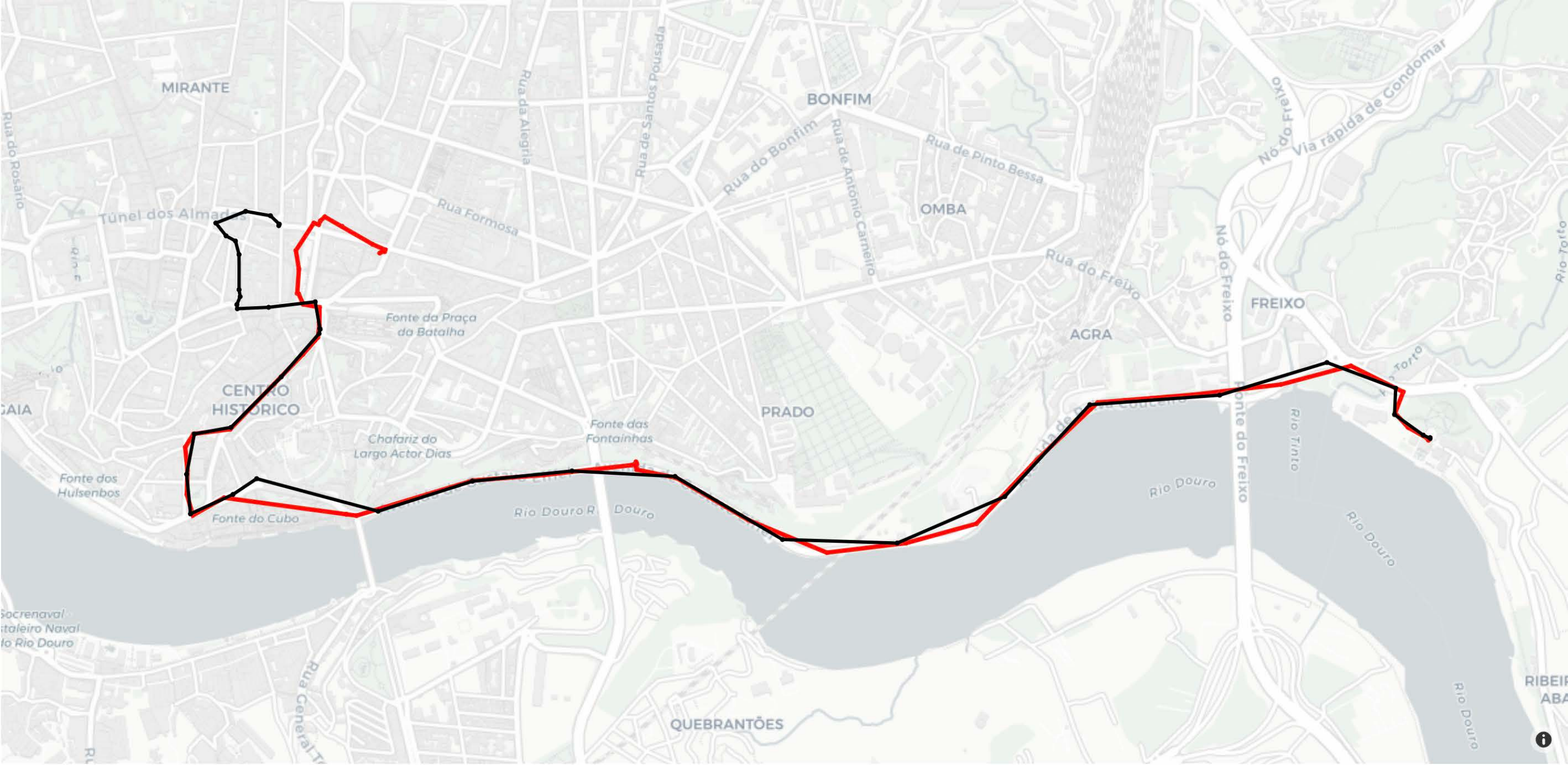}
    \includegraphics[width=2cm,height=1.5cm]{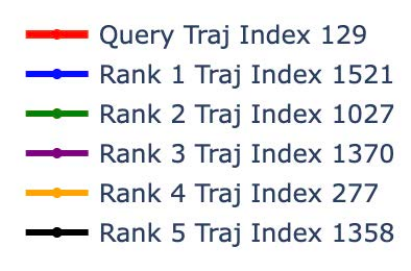}
    }

    \subfloat[TrajCL Visualizations]{
    \label{subfig:trajcl2}
    \includegraphics[width=3cm,height=2cm]{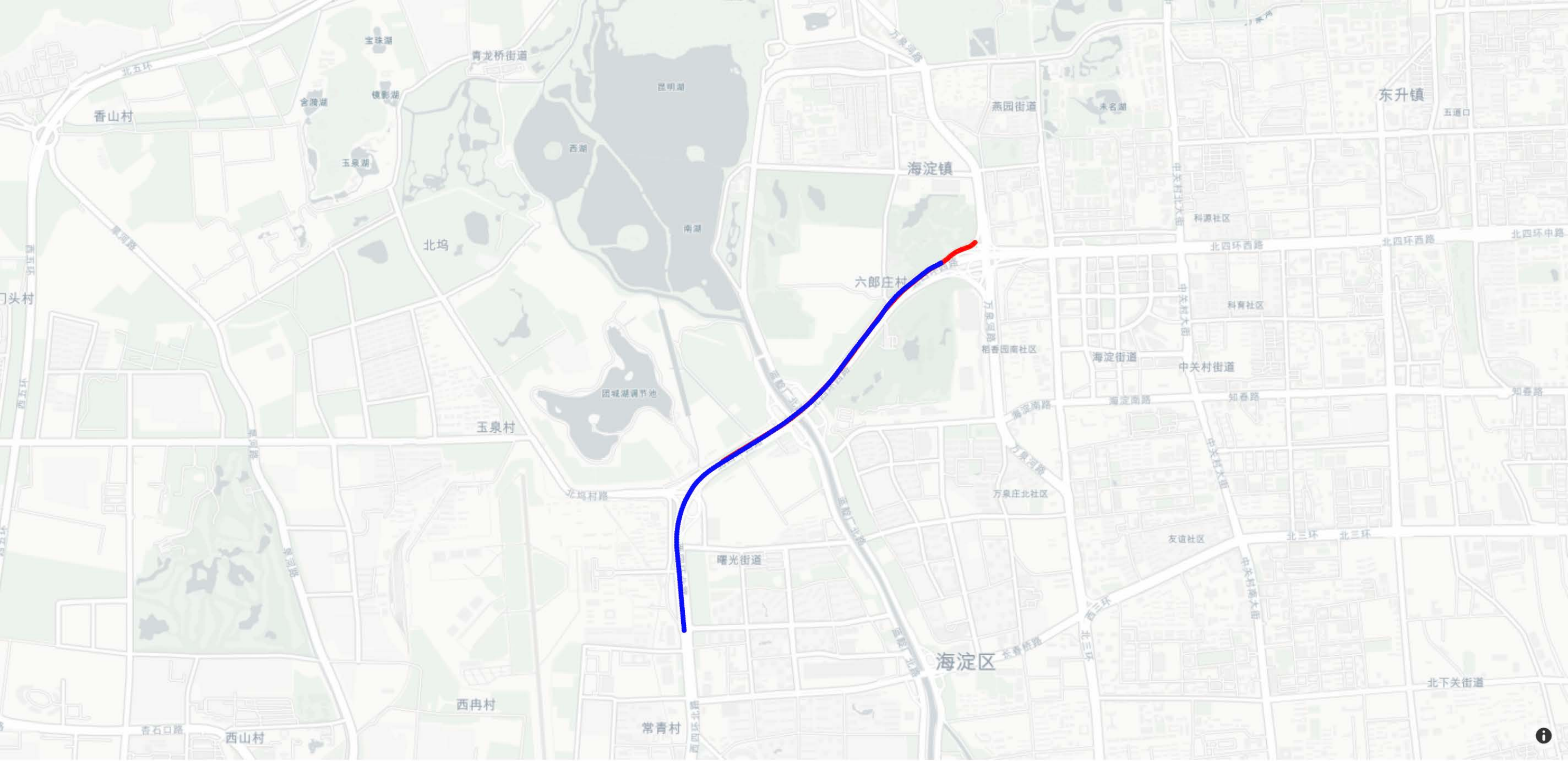}
    \includegraphics[width=3cm,height=2cm]{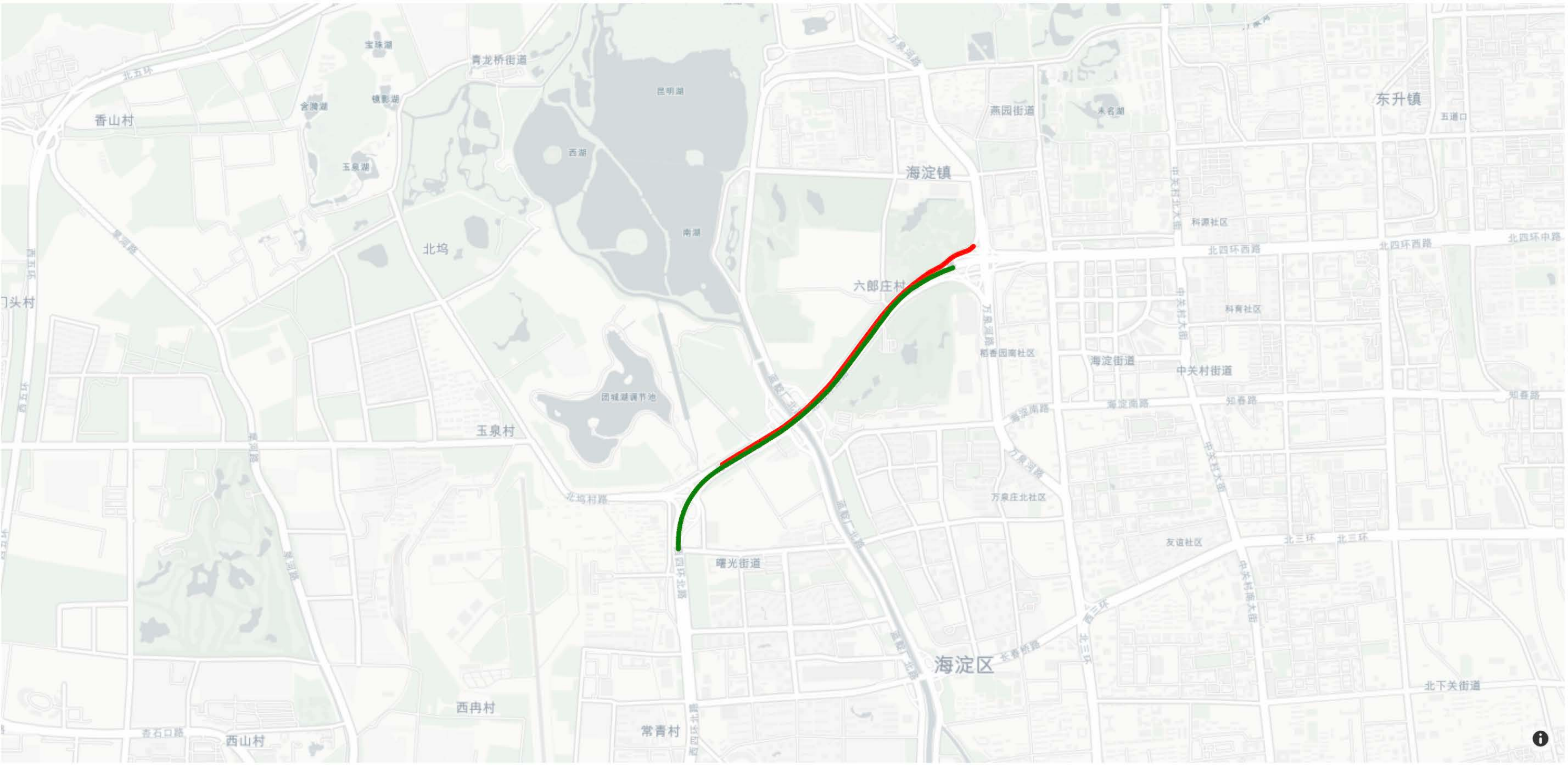}
    \includegraphics[width=3cm,height=2cm]{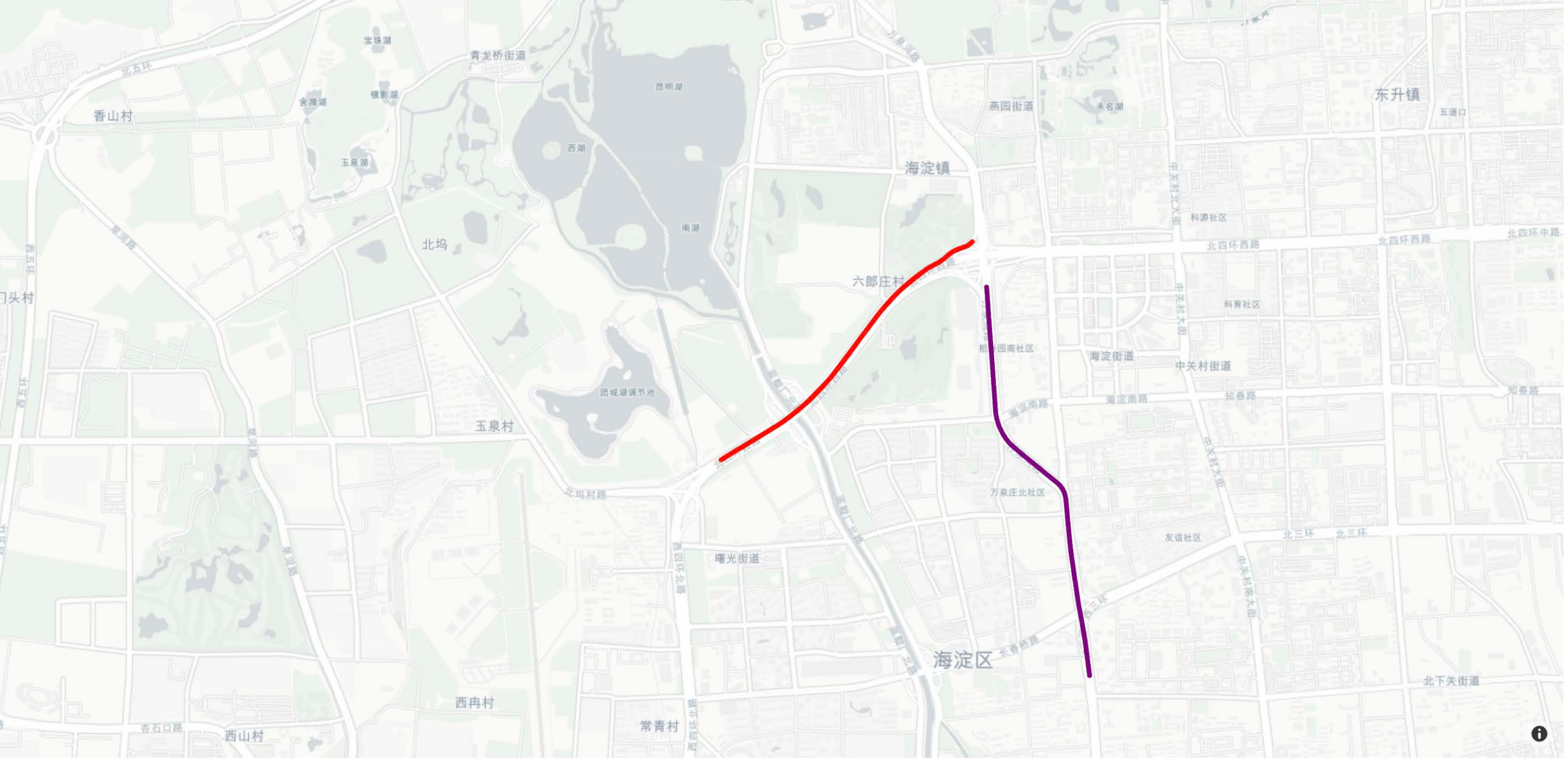}
    \includegraphics[width=3cm,height=2cm]{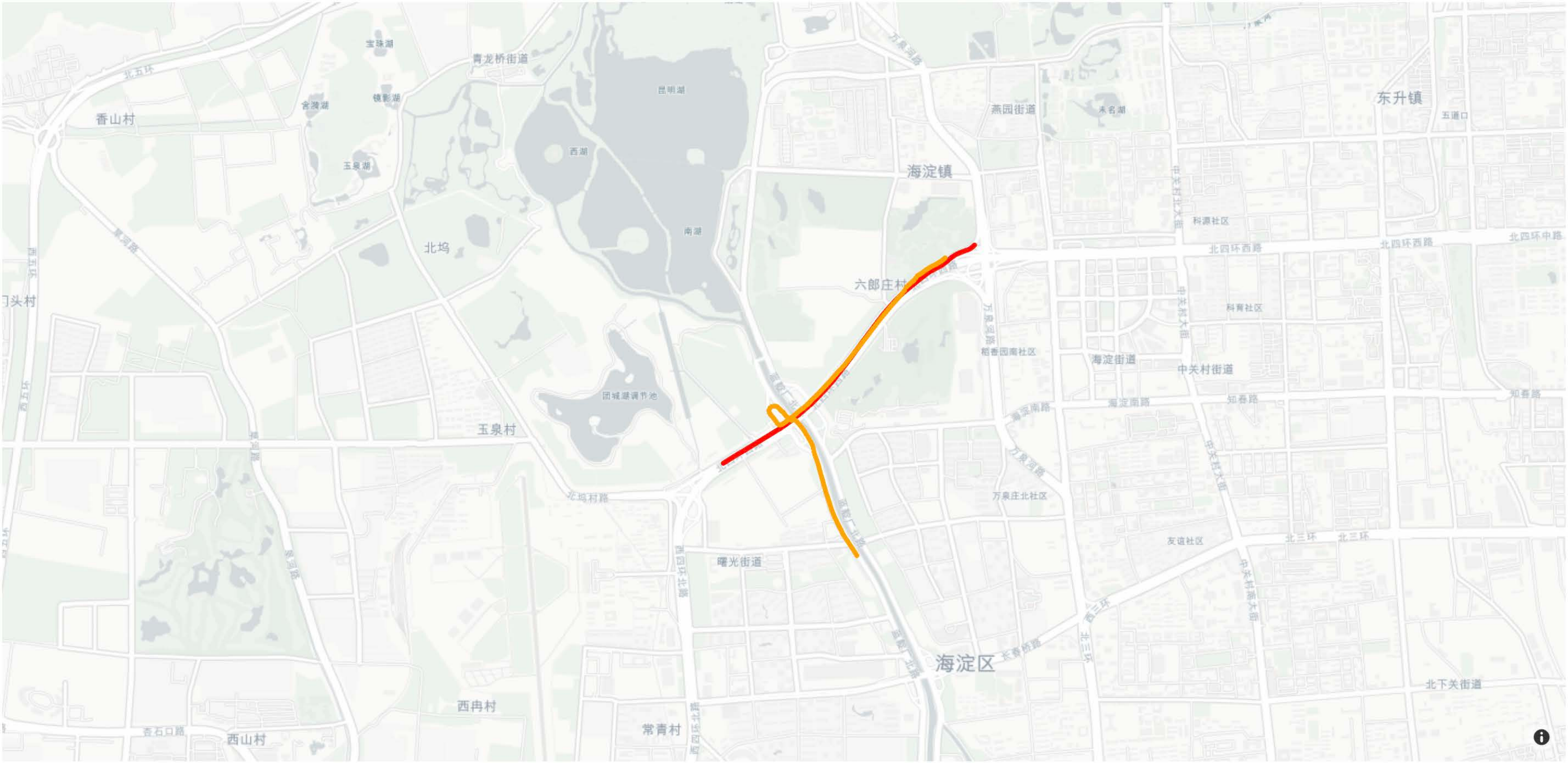}
    \includegraphics[width=3cm,height=2cm]{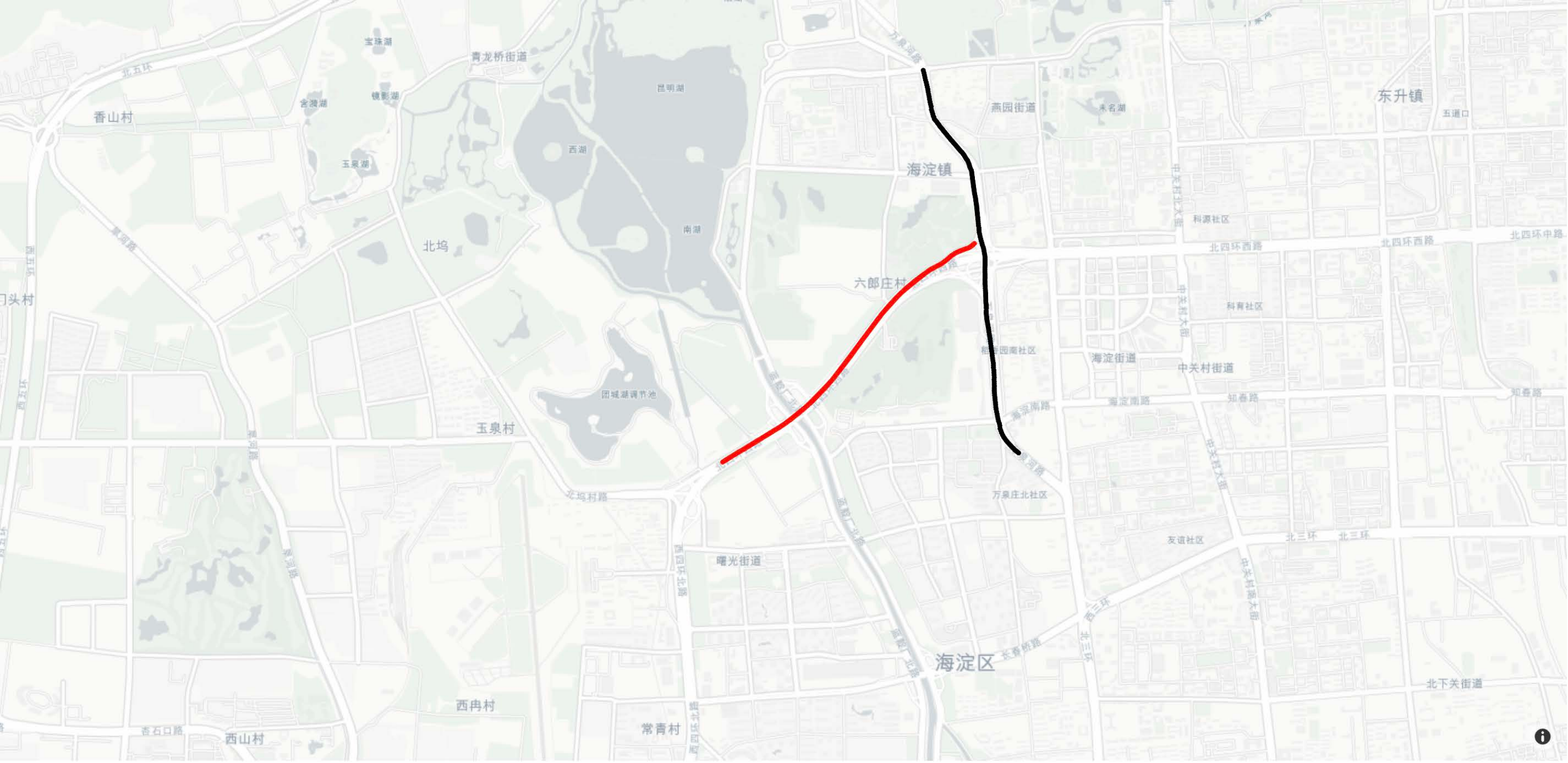}
    \includegraphics[width=2cm,height=1.5cm]{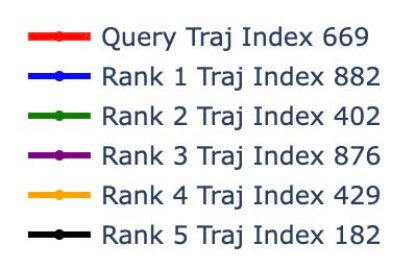}
    }
    
    \subfloat[T-JEPA Visualizations]{
    \label{subfig:tjepa2}
    \includegraphics[width=3cm,height=2cm]{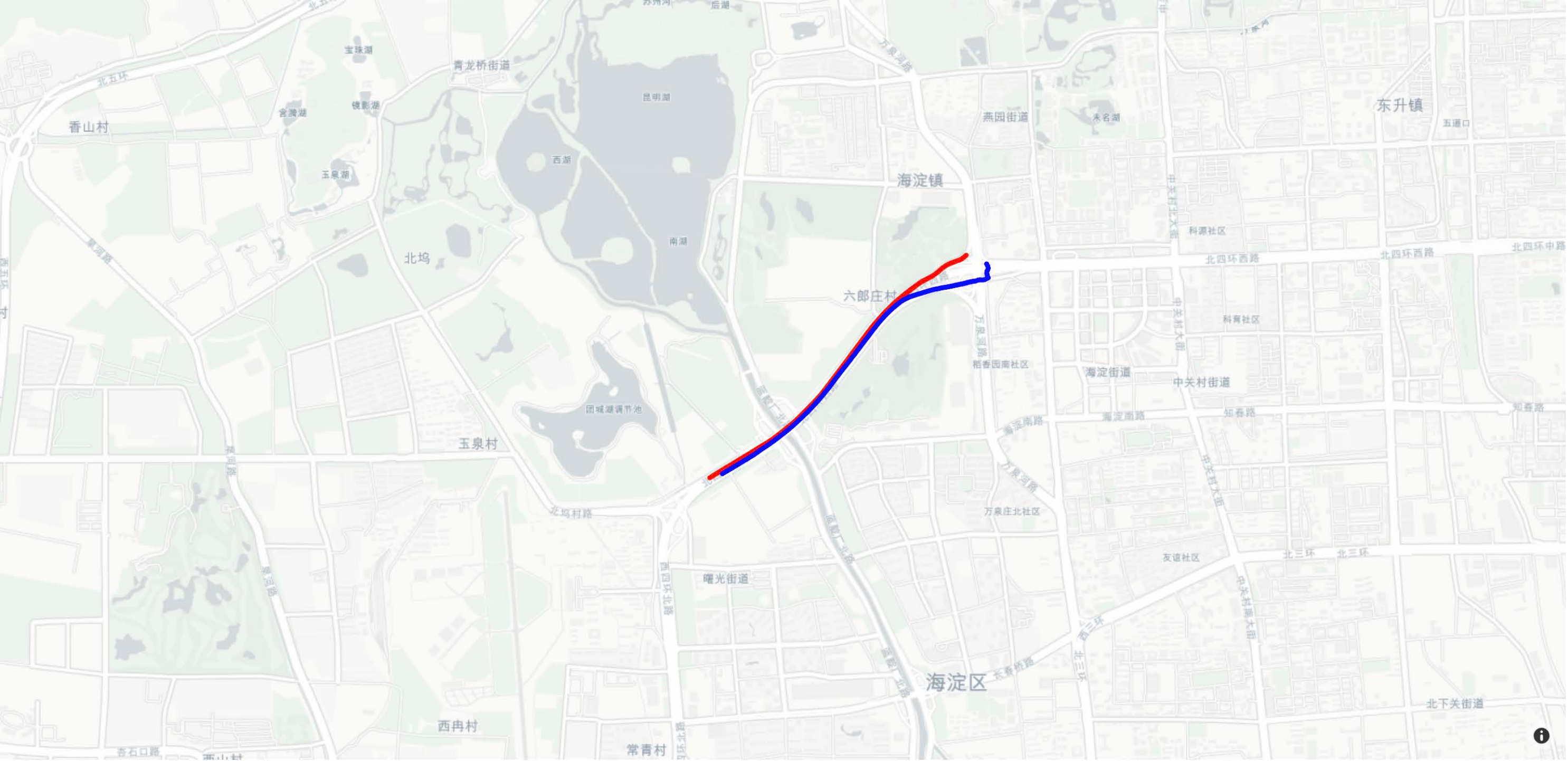}
    \includegraphics[width=3cm,height=2cm]{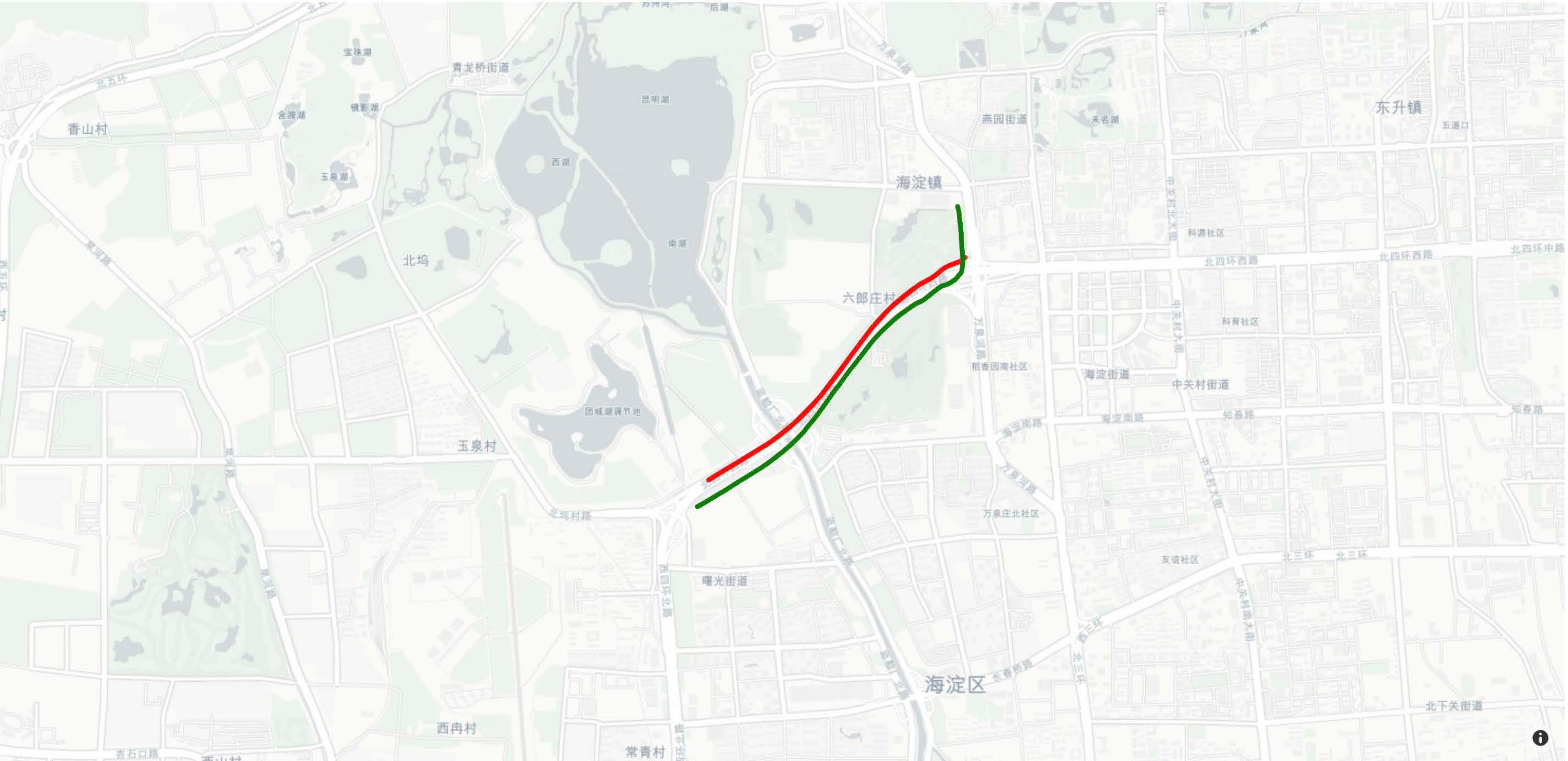}
    \includegraphics[width=3cm,height=2cm]{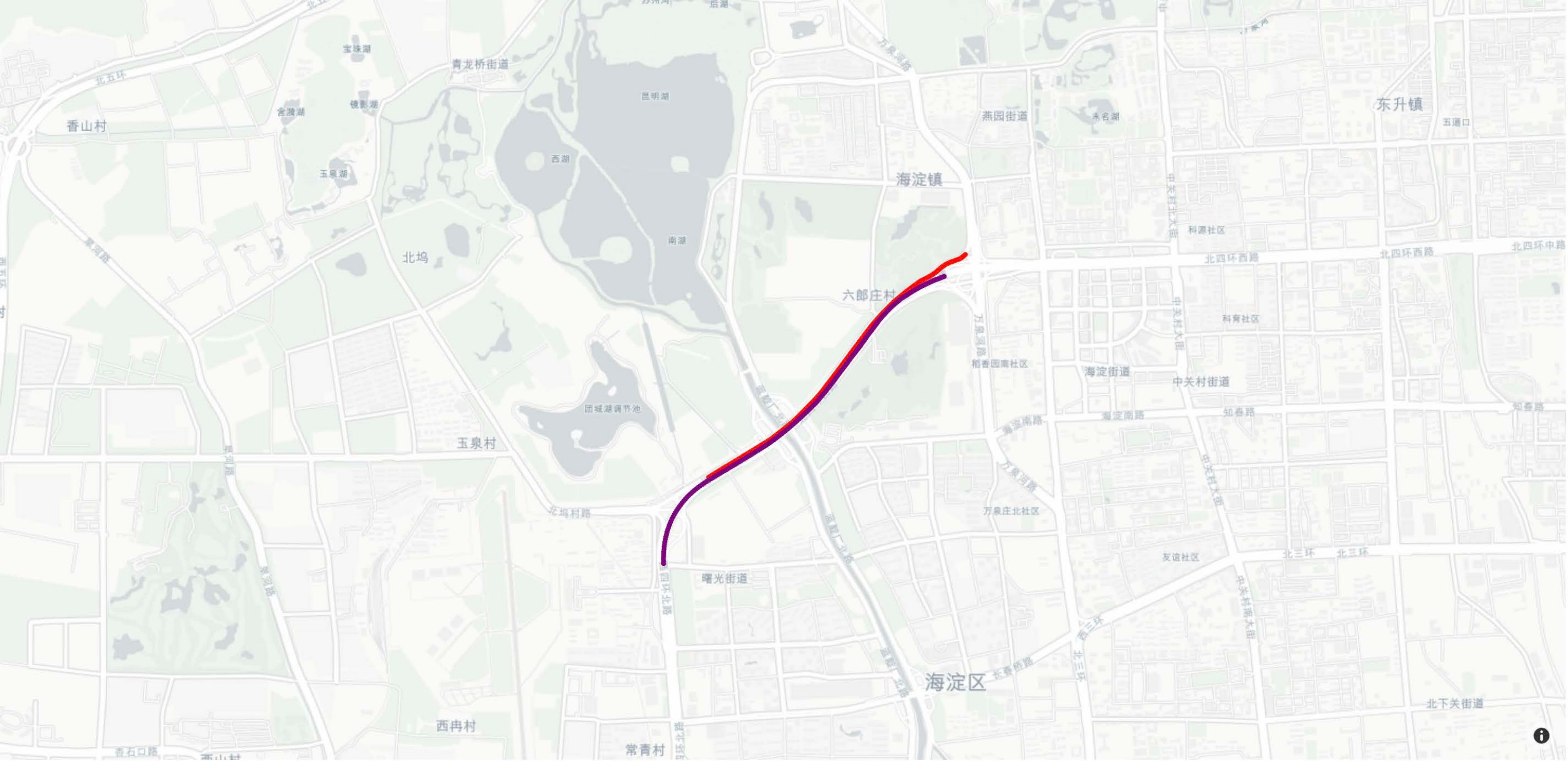}
    \includegraphics[width=3cm,height=2cm]{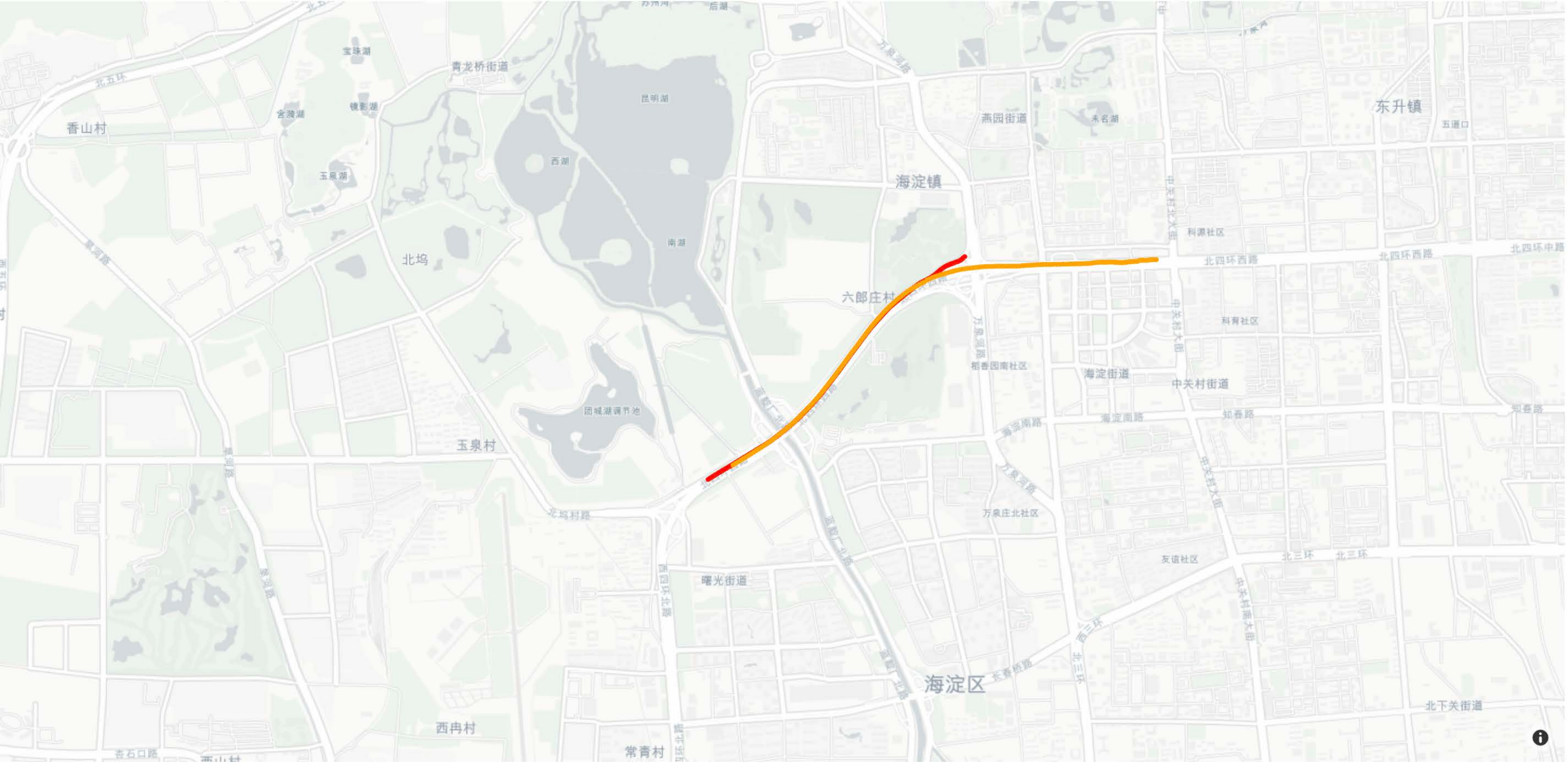}
    \includegraphics[width=3cm,height=2cm]{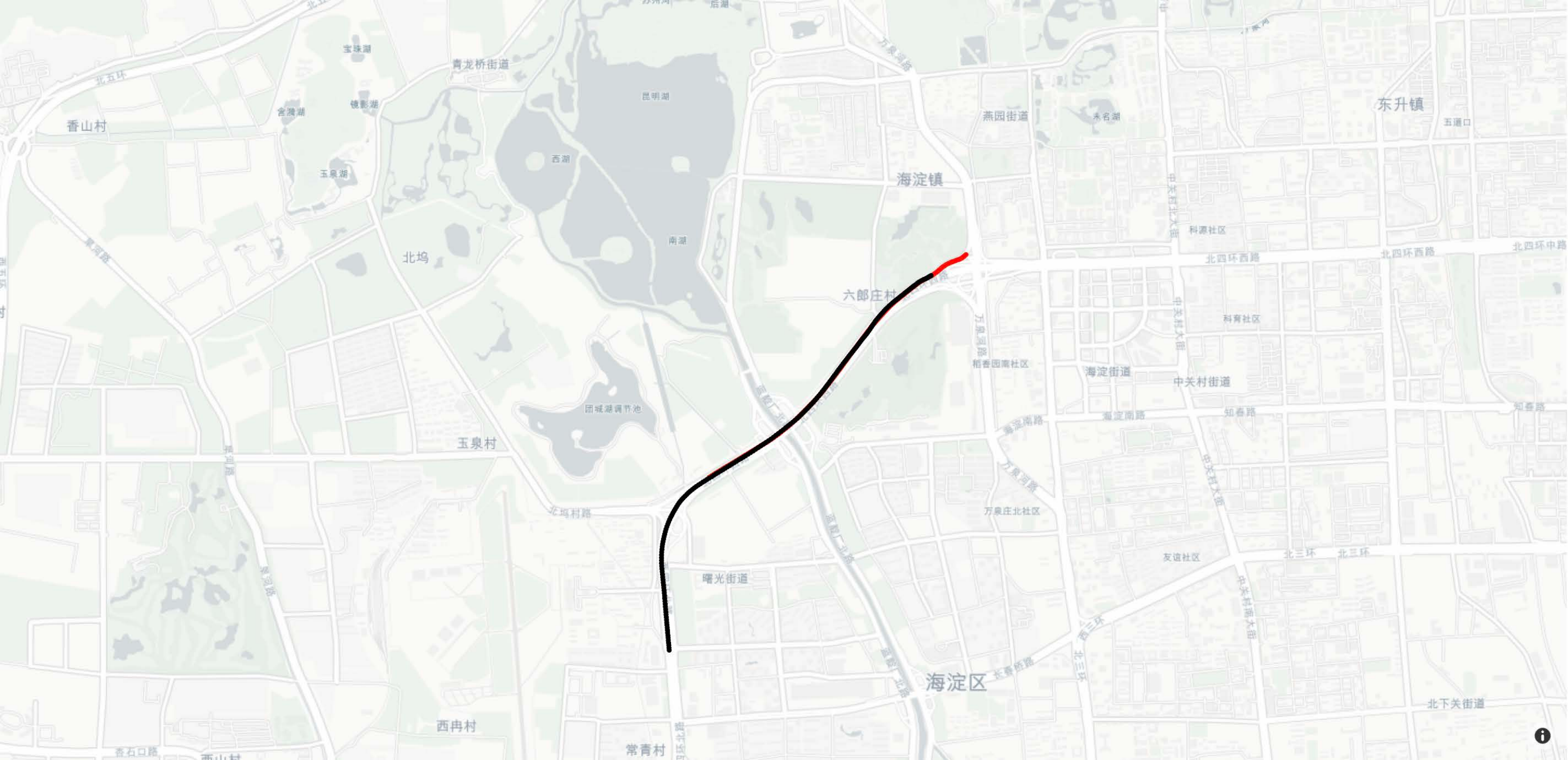}
    \includegraphics[width=2cm,height=1.5cm]{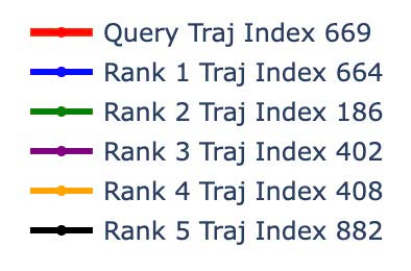}
    }
    
    \caption{Visual comparisons between TrajCL and T-JEPA on 5-NN query after being fine-tuned in Hausdorff measure. The first set of comparisons (a) and (b) is in Porto and the second set (c) and (d) is in GeoLife.}
    \label{fig:qualitative}
\end{figure*}

\begin{figure*}[htbp]
    \centering
    \subfloat{
    \includegraphics[scale=0.1]{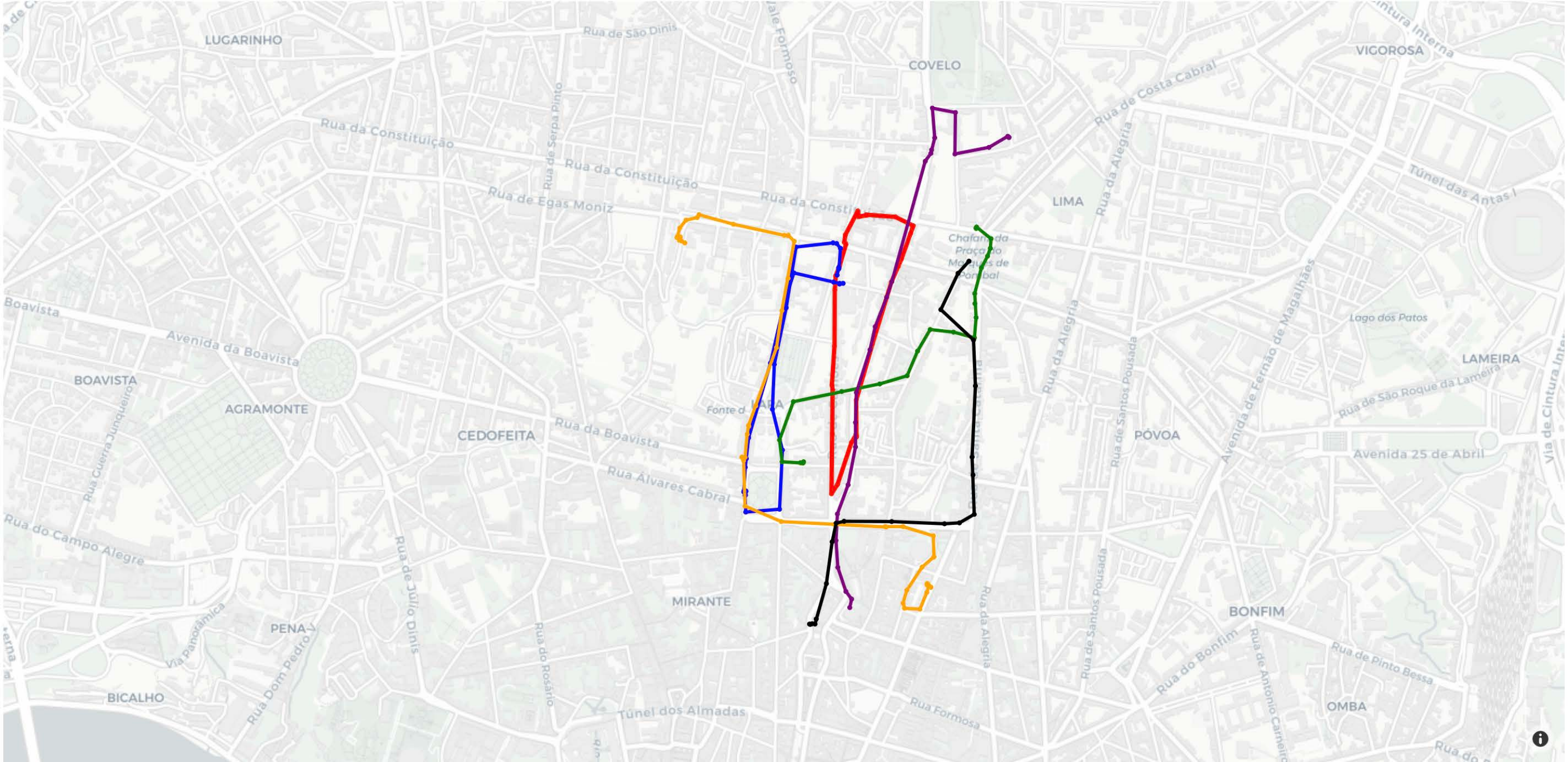}
    \includegraphics[scale=0.1]{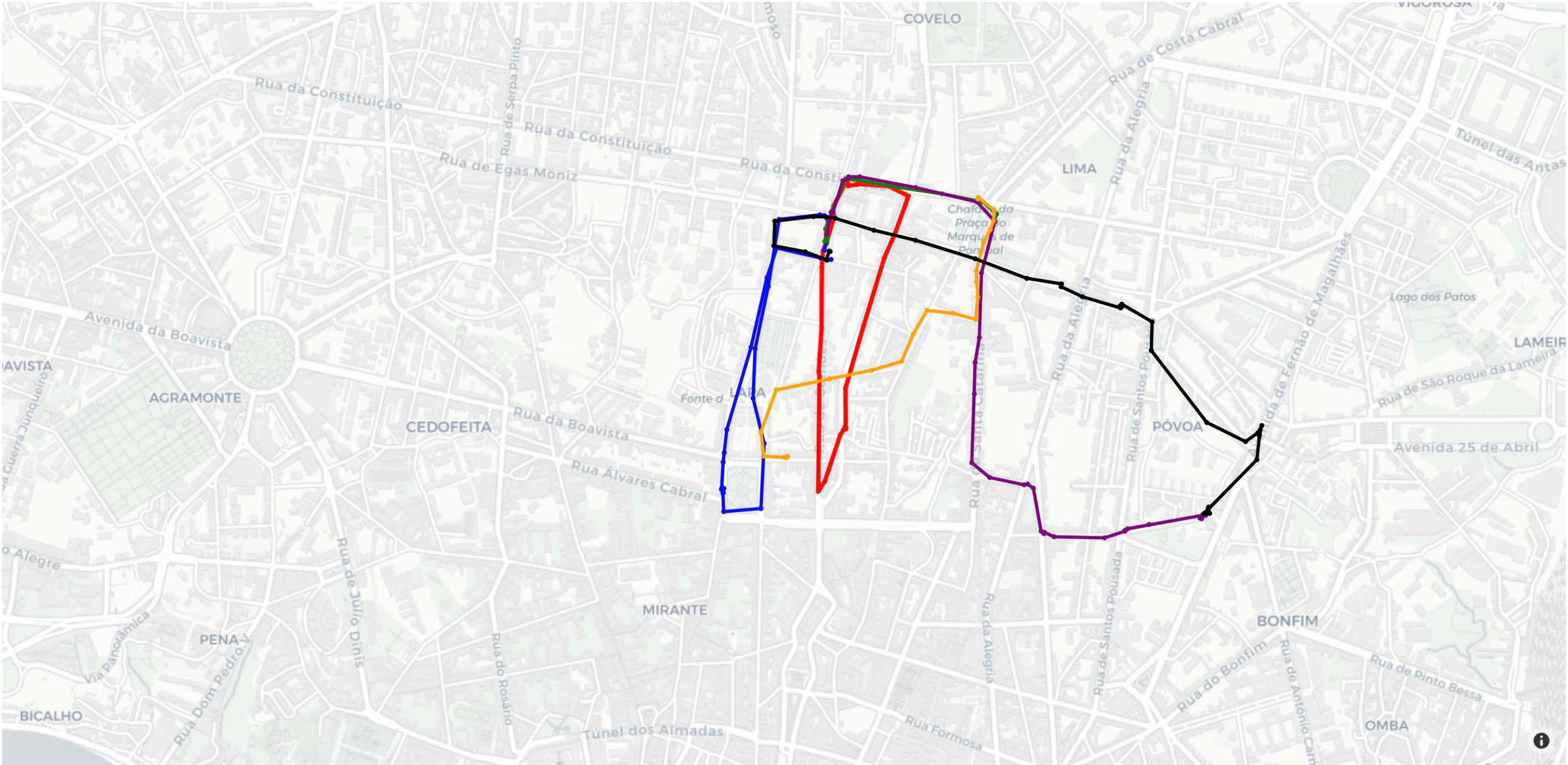}
    \includegraphics[scale=0.1]{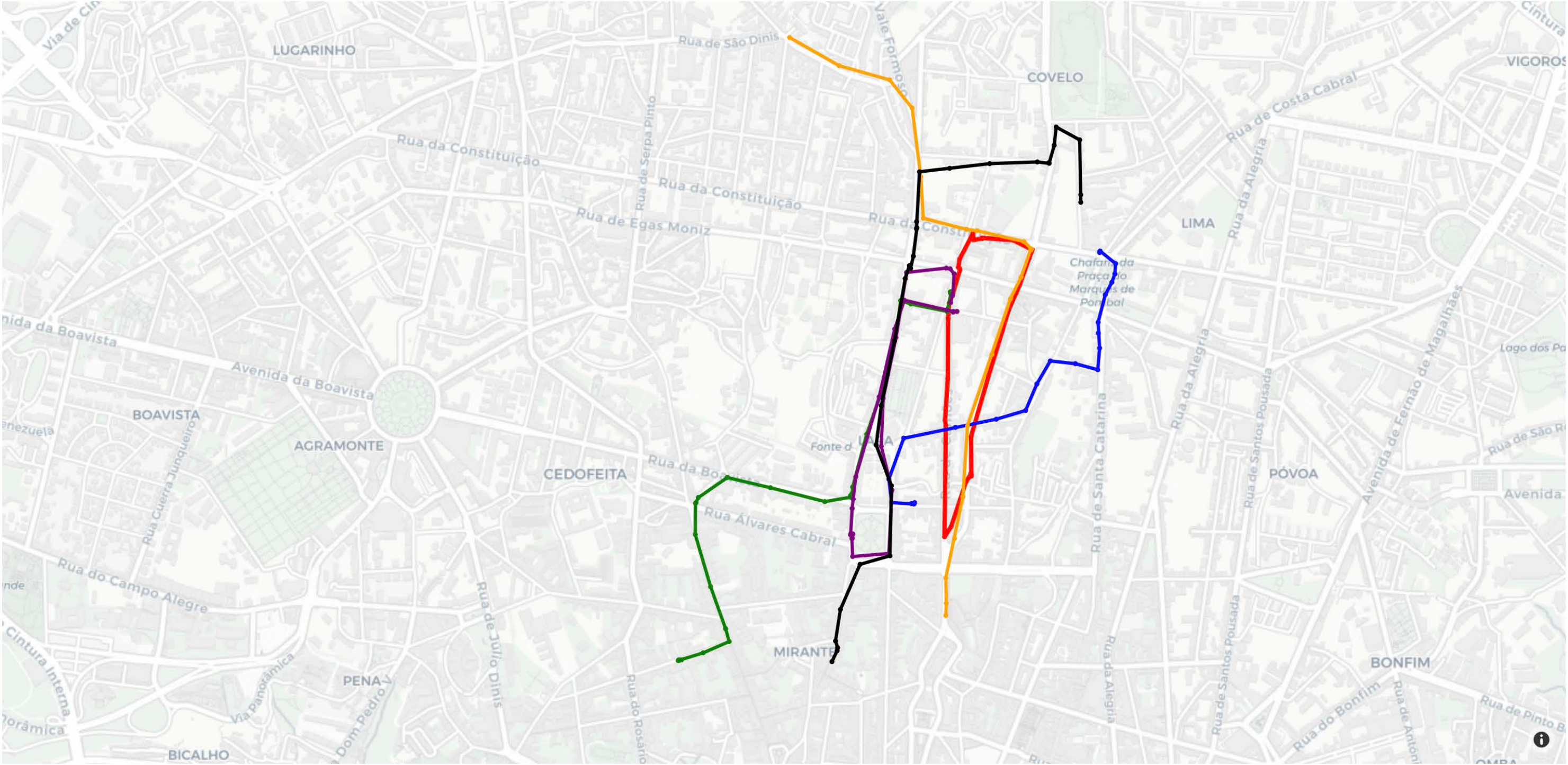}
    }

    \caption{A challenging case where all 3 models have limited performances: t2vec (left), TrajCL (middle), T-JEPA (right). This case is sampled from Porto.}
    \label{fig:limit}
\end{figure*}

Table \ref{tab:finetune} demonstrates the results by t2vec, TrajCL, and T-JEPA where the higher values mean better results. We find that after freezing the encoder, the extracted representations by t2vec struggle to approximate any heuristic measures, where the hit ratios and recalls are significantly lower than the results reported in \cite{chang2023contrastive} by tuning the last encoder layer. Compared with TrajCL, our T-JEPA performs the highest HR@5, HR@20, and R5@20 in Porto and GeoLife for all measures, except for HR@20 in Fréchet. Even though TrajCL outperforms T-JEPA in Porto in previous experiments, but results in an average value 4.1\% lower than us. This implies our T-JEPA has better generalization potential. For results in T-Drive, although TrajCL performs better in Hausdoff and Fréchet measures, the averaged overall result is still 1.1\% lower than T-JEPA. This reflects we perform even better on EDR and LCSS.

The Superiority of the fine-tuning tasks verifies that T-JEPA has stronger generalization abilities which allows it to adapt to multiple heuristic measures. It is worth mentioning that T-JEPA requires no manual data augmentation schemes compared to TrajCL, which relies heavily on pre-defined augmentation methods. This automation not only simplifies the training process but also enhances the generalized and robust representation learning from only raw data.

\subsection{Qualitative Evaluation}

We have also visualized some cases where T-JEPA outperforms TrajCL on approximation to heuristic measures. Fig. \ref{fig:qualitative} demonstrates 2 sets of comparisons of 5-NN queries between TrajCL and T-JEPA after fine-tuning by the Hausdorff measure. Each row represents the rank 1 to 5 matched trajectories from left to right by the red query trajectories. The rightmost figures provide the corresponding indexes of the query and the matched trajectories. We can observe that T-JEPA can successfully find more similar trajectories, especially for ranks 3--5. The more similar matched trajectories from visualizations provide a highly valuable reference for similarity computation accuracy, confirming the generalized and robust learned representations by T-JEPA.

% \begin{figure}[ht]
%     \centering
%     \subfloat[no AdjFuse.]
%     {\includegraphics[width=0.23\textwidth]{no_adj.pdf}
%     \label{fig:no_adj}
%     }
%     \hfill
%     \subfloat[lower/higher sampling ratios.]
%     {\includegraphics[width=0.23\textwidth]{lo_hi.pdf}
%     \label{fig:lo_hi}
%     }
%     % \setlength{\abovecaptionskip}{-0.2cm}
%     % \setlength{\belowcaptionskip}{-0.3cm}
%     \caption{Ablation studies.}
%     \label{fig:ablations}
% \end{figure}

\begin{figure}[ht]
    \centering
    \begin{subfigure}[b]{0.23\textwidth}
        \centering
        \includegraphics[width=\textwidth]{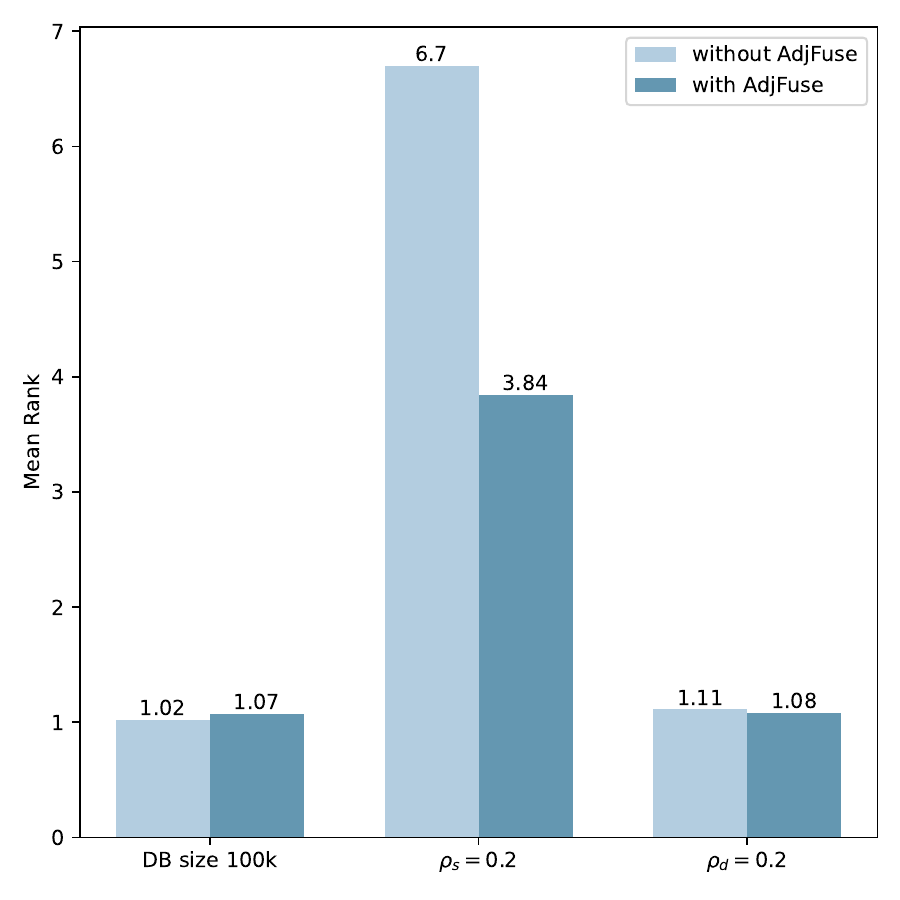}
        \caption{No AdjFuse.}
        \label{fig:no_adj}
    \end{subfigure}
    \hfill
    \begin{subfigure}[b]{0.23\textwidth}
        \centering
        \includegraphics[width=\textwidth]{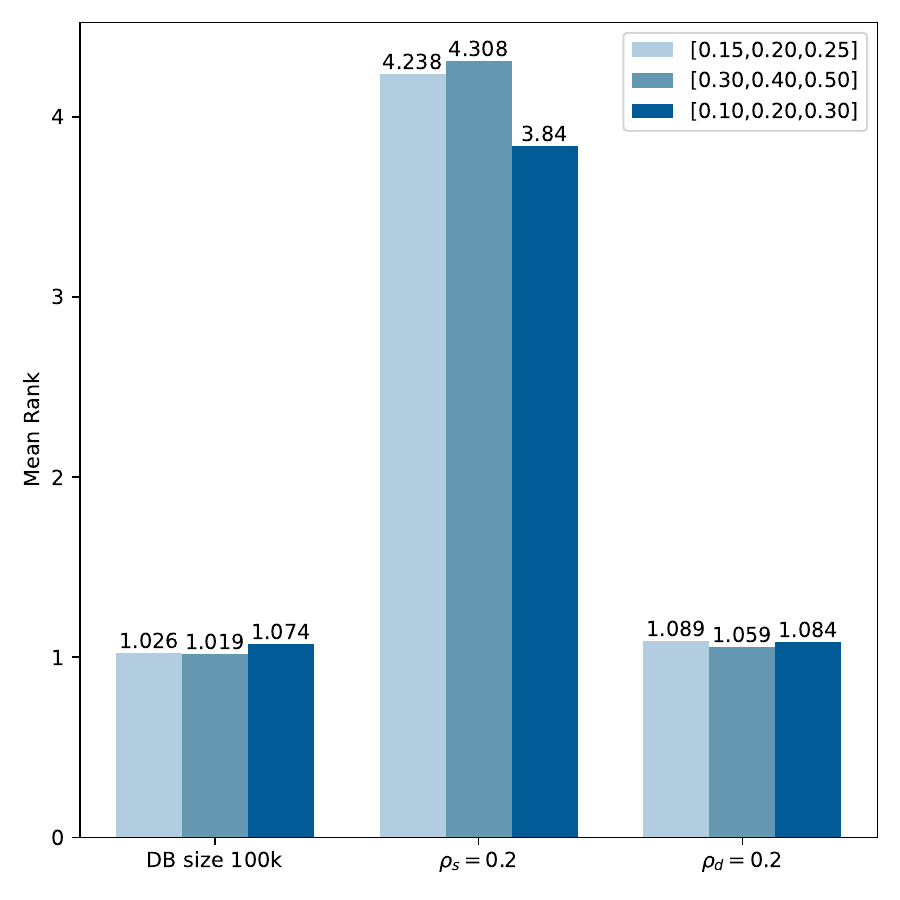}
        \caption{Lower/Higher target ratios.}
        \label{fig:lo_hi}
    \end{subfigure}
    \caption{Ablation studies.}
    \label{fig:ablations}
\end{figure}
% \begin{figure}[ht]
%     \centering
%     \begin{minipage}[b]{0.23\textwidth}
%         \centering
%         \includegraphics[width=\textwidth]{no_adj.pdf}
%         \captionof{subfig}{No AdjFuse.}
%         \label{fig:no_adj}
%     \end{minipage}
%     \hfill
%     \begin{minipage}[b]{0.23\textwidth}
%         \centering
%         \includegraphics[width=\textwidth]{lo_hi.pdf}
%         \captionof{subfig}{Lower/Higher sampling ratios.}
%         \label{fig:lo_hi}
%     \end{minipage}
%     \caption{Ablation studies.}
%     \label{fig:ablations}
% \end{figure}

\subsection{Ablation study}
We apply ablation studies to show the efficacy of our proposed components. \\
\textbf{Influence of no AdjFuse module.} We remove the $AdjFuse$ module and compare it with T-JEPA on the most similar trajectory search performance. In Fig. \ref{fig:no_adj}, we provide results on 100k database size, down-sampling rate $\rho_{s}=0.2$ and distortion rate $\rho_{d}=0.2$. We can find that without the $AdjFuse$ module, even though the mean rank in 100k database size has a slight improvement of 0.05, the performance in the down-sampling case slumps to 6.7 which is 74\% worse than T-JEPA with $AdjFuse$. The performance in the distortion case also drops 0.03. Therefore, it is easy to see that our proposed $AdjFuse$ produces more robust results, especially in down-sampled trajectories with low and irregular sampling rates.

\noindent\textbf{Influence of lower and higher targets sampling ratios.} To verify the impact of different target sampling rates on T-JEPA, we take an extra 2 sets of sampling ratios \{0.05, 0.15, 0.25\}  and \{0.30, 0.40, 0.50\}. We also compare them with T-JEPA on the most similar trajectory search performance. As shown in Fig. \ref{fig:lo_hi}, we can see that higher sampling ratios boost the performance in searching in the 100k database by 0.055, and in 
$\rho_{d}=0.2$ distortion rate by 0.025. This is due to more challenging samples with less context information being given to the model when training as higher target sampling ratios indicate less sampled context after overlap removal. The lower sampling ratios produce average performance in these two cases. However, both lower and higher sampling ratios suffer from the down-sampling case, where the performances are dropped by 10.36\% and 12.19\%. Therefore, sampling rates in \{0.10, 0.20, 0.30\} produce the most robust results by providing the model with proper training difficulties to ensure effective learning.

\section{Limits}
We include a case study where all of the t2vec, TrajCL, and our T-JEPA fail to produce a high HR@5 for the Hausdorff measure. We can find from Fig. \ref{fig:limit} that the query trajectory in red forms the shape of a closed loop and the matched 5-NN trajectories exhibit very inconsistent patterns. This is a challenging case that closed-looped trajectories usually involve repetitive patterns, where spatial overlaps further increase the confusion in capturing their spatio-temporal features. Therefore, it complicates the accurate modeling of such trajectories. Future solutions may include enriching behavioral information for trajectory modeling, such as point-of-interests and road networks. And effective geospatial knowledge fusion would significantly improve the trajectory modeling accuracy.

\section{Conclusion}
In this paper, we introduced T-JEPA, a trajectory similarity computation model leveraging the Joint Embedding Predictive Architecture to automate data augmentation and enhance high-level semantic understanding. The automatic resampling and unique predictive processes enable more robust and generalized trajectory representations. Additionally, the $AdjFuse$ module enriches spatial contextual information, improving the stability of representation learning on low and irregularly sampled data. Our experiments on GPS and FourSquare datasets demonstrate that T-JEPA achieves overall superior performance in trajectory similarity computation. This provides a novel solution for urban trajectory representation learning and advances the field of trajectory similarity computation.
%%
%% The acknowledgments section is defined using the "acks" environment
%% (and NOT an unnumbered section). This ensures the proper
%% identification of the section in the article metadata, and the
%% consistent spelling of the heading.
\begin{acks}
We would like to acknowledge the support of the Cisco’s National Industry Innovation Network (NIIN) Research Chair Program and the ARC Centre of Excellence for Automated Decision-Making and Society (CE200100005). The research utilized computing resources and services provided by Gadi, supercomputer of the National Computational Infrastructure (NCI) supported by the Australian Government.
\end{acks}

% \newpage
%%
%% The next two lines define the bibliography style to be used, and
%% the bibliography file.
\bibliographystyle{ACM-Reference-Format}
\bibliography{sample-base}

%%
%% If your work has an appendix, this is the place to put it.
% \appendix

% \section{Research Methods}

% \subsection{Part One}

% Lorem ipsum dolor sit amet, consectetur adipiscing elit. Morbi
% malesuada, quam in pulvinar varius, metus nunc fermentum urna, id
% sollicitudin purus odio sit amet enim. Aliquam ullamcorper eu ipsum
% vel mollis. Curabitur quis dictum nisl. Phasellus vel semper risus, et
% lacinia dolor. Integer ultricies commodo sem nec semper.

% \subsection{Part Two}

% Etiam commodo feugiat nisl pulvinar pellentesque. Etiam auctor sodales
% ligula, non varius nibh pulvinar semper. Suspendisse nec lectus non
% ipsum convallis congue hendrerit vitae sapien. Donec at laoreet
% eros. Vivamus non purus placerat, scelerisque diam eu, cursus
% ante. Etiam aliquam tortor auctor efficitur mattis.

% \section{Online Resources}

% Nam id fermentum dui. Suspendisse sagittis tortor a nulla mollis, in
% pulvinar ex pretium. Sed interdum orci quis metus euismod, et sagittis
% enim maximus. Vestibulum gravida massa ut felis suscipit
% congue. Quisque mattis elit a risus ultrices commodo venenatis eget
% dui. Etiam sagittis eleifend elementum.

% Nam interdum magna at lectus dignissim, ac dignissim lorem
% rhoncus. Maecenas eu arcu ac neque placerat aliquam. Nunc pulvinar
% massa et mattis lacinia.

\end{document}